\RequirePackage{fix-cm}
\documentclass[twocolumn]{svjour3}          
\smartqed  
\usepackage{graphicx}
\usepackage{natbib}
\usepackage{xcolor}
\usepackage{appendix}
\usepackage{multirow}
\usepackage{multicol}
\usepackage{stfloats}
\usepackage{enumitem}
\usepackage{amsmath}
\usepackage{amsfonts}
\usepackage{booktabs}
\usepackage{subfigure}
\usepackage{graphics}
\usepackage{animate}
\usepackage{bm}
\usepackage{mathrsfs}
\usepackage{verbatim}

\newcommand{\DN}{HumanAct12}
\newcommand{\PDM}{pose decoding module}
\newcommand{\revision}[1]{\textcolor{black}{#1}}
\newcommand{\revisions}[1]{\textcolor{black}{#1}}

\newcommand{\et}[2]{${#1}^{\pm{#2}}$}
\newcommand{\etb}[2]{$\mathbf{{#1}}^{\pm{#2}}$}
\newcommand{\ets}[2]{$\underline{{#1}}^{\pm{#2}}$}
\begin{document}

\title{Action2video: Generating Videos of Human 3D Actions
}


\author{Chuan Guo         \and
        Xinxin Zuo \and
        Sen Wang \and
        Xinshuang Liu \and 
        Shihao Zou \and 
        Minglun Gong \and 
        Li Cheng
}



\institute{Chuan Guo, Xinxin Zuo, Sen Wang, Shihao Zou, Li Cheng are with the Department of Electrical and Computer Engineering, University of Alberta. E-mails: \{cguo2, xzuo, sen9, szou2, lcheng5\}@ualberta.ca.    \\       
           Xinshuang Liu is with the School of Software, Tsinghua University, Beijing 100084, China. E-mail: liuxs17@mails.tsinghua.edu.cn. \\
           Minglun Gong is with the School of Computer Science, University of Guelph. E-mail: minglun@uoguelph.ca.
}

\date{Received: date / Accepted: date}

\maketitle
\begin{figure*}[t]
  \centering
  \includegraphics[width=0.9\linewidth]{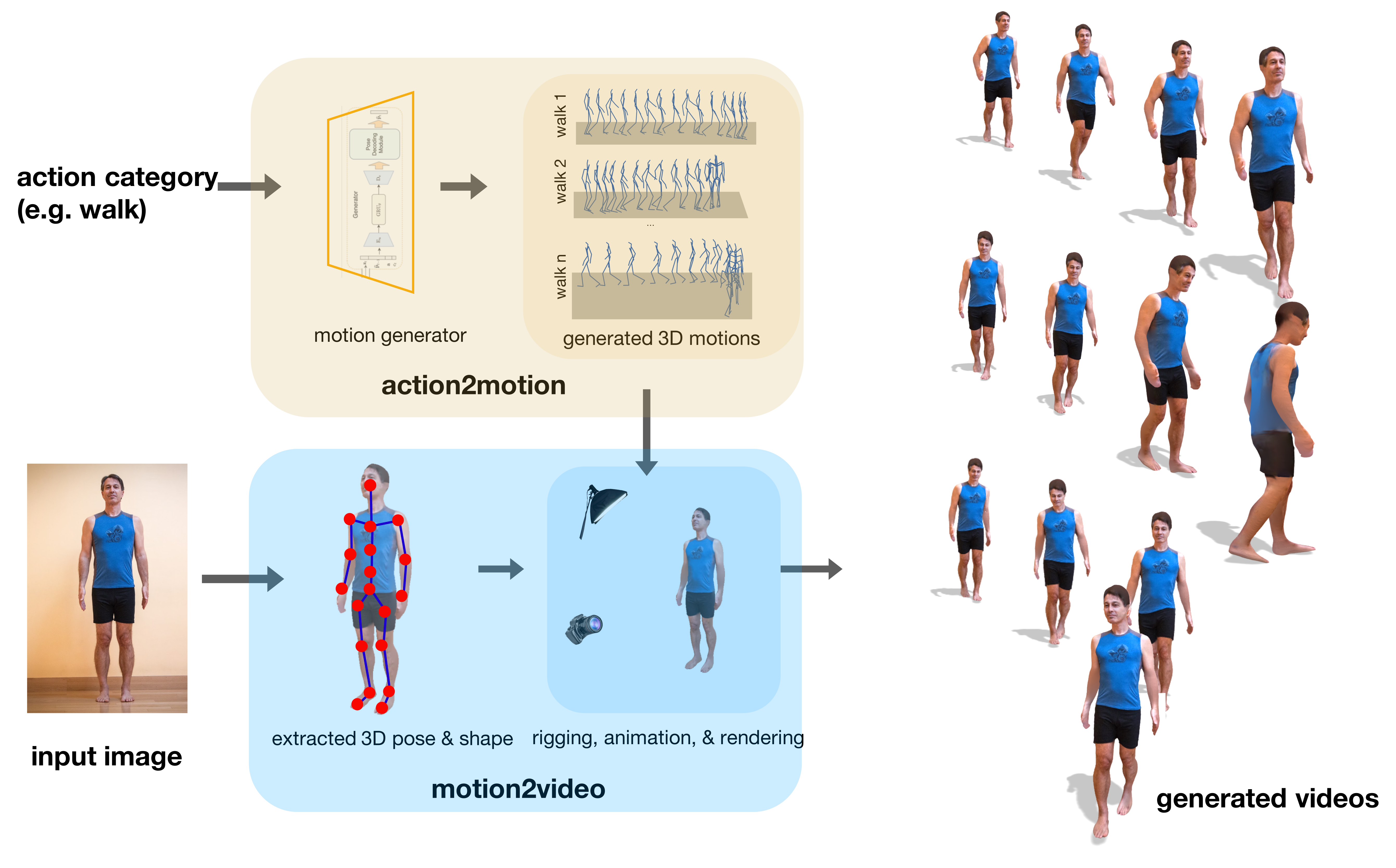}
  \caption{Our action2video pipeline generates human full-body motion videos of prescribed actions in two steps: \textit{action2motion} first generates diverse and natural 3D motions of predefined actions; \textit{motion2video} proceeds to extract 3D surface shape and texture from an additional 2D input image, and to render 2D videos of the generated motions.}
  \label{fig:teaser}
  \vspace{-0.3cm}
\end{figure*}

\begin{abstract}
We aim to tackle the interesting yet challenging problem of generating videos of \textit{diverse} and \textit{natural} human motions from prescribed action categories. 
The key issue lies in the ability to synthesize multiple distinct motion sequences that are realistic in their visual appearances.
It is achieved in this paper by a two-step process that maintains internal 3D pose and shape representations, \textit{action2motion} and \textit{motion2video}. Action2motion stochastically generates plausible 3D pose sequences of a prescribed action category, which are processed and rendered by motion2video to form 2D videos. 
Specifically, the Lie algebraic theory is engaged in representing \textit{natural} human motions following the physical law of human kinematics; 
a temporal variational auto-encoder (VAE) is developed that encourages \textit{diversity} of output motions. 
Moreover, 
given an additional input image of a clothed human character, an entire pipeline is proposed to extract his/her 3D detailed shape, and to render in videos the plausible motions from different views.   
This is realized by improving existing methods to extract 3D human shapes and textures from single 2D images, rigging, animating, and rendering to form 2D videos of human motions.
It also necessitates the curation and reannotation of 3D human motion datasets for training purpose.
Thorough empirical experiments including ablation study, qualitative and quantitative evaluations manifest the applicability of our approach, and demonstrate its competitiveness in addressing related tasks, where components of our approach are compared favorably to the state-of-the-arts.

\end{abstract}

\section{Introduction}
\label{intro}
Human-centric activities always play a key role in our daily life. 
In recent years, noticeable progresses have been made in video forecasting~\citep{wu2020future,gao2019disentangling} and synthesis~\citep{zhu2020s3vae,tulyakov2018mocogan,vondrick2017generating,denton2018stochastic}. Meanwhile, it remains a substantial challenge in generating realistic videos of diverse and plausible human motions. This is evidenced in many recent 
video generation efforts~\citep{yang2018pose,cai2018deep,kim2019unsupervised}, where the appearances of synthesized human characters are unfortunately either blurring or surreal, and are still far from being photo-realistic; their motions are often distorted and unnatural. These observations stress the importance of properly modeling 
human body postures \& temporal articulations, as well as the surface shapes and textures of the local body parts. It also motivates us to examine the problem of generating videos of human motions based on action categories, the basic ingredient of human behaviors.

Due to the complexity of human articulations and pose dynamics, generating human videos is far from being trivial. 
Existing efforts usually represent human motions in 2D space, which are then rendered pixel-wise to form 2D videos. 
Moreover, extra information such as an initial 2D pose or a partial/entire motion sequence is usually required, which is practically undesirable. For instance, \cite{yang2018pose} produces deterministic sequence of 2D motions, 
which is followed by synthesizing the appearances frame-by-frame through adversarial training. 
Action-conditioned 2D human behavior modeling is also studied in~\cite{cai2018deep}, where 2D pose generator and motion generator are trained progressively. 
Very recently, \revision{the efforts of~\citep{weng2019photo,huang2020arch} consider the related task of extracting 3D characters from single images, which is then animated to form 3D motions;} \cite{de2019generating} addresses another related task of generating human action videos by composing the human motions and scenes with probabilistic graphical models in 3D game engine;. However, the motions used in both methods are real-life motions that have been made available in prior, instead of being synthesized on the spot.

Overall, the existing methods fall short in the following aspects: 1) direct modeling of 2D motions is inherently insufficient to capture the underlying 3D human pose articulations and shape deformations. The absence of 3D geometric information often leads to visual distortions and ambiguities; 2) coordinate locations of body joints are commonly used as the human pose representation, which undesirably entangle the human skeletons and their motion trajectories. Moreover, this creates extra barriers in modeling human kinematics; 3) initial \revision{poses} often impede the diversity of generated human dynamics. For example, in actions such as \textit{warm up} and \textit{boxing}, initial poses crucially influence the formation of the rest sequences; and 4) \revision{the popular choice of pixel-to-pixel synthesis among existing efforts on action conditioned video generation has been evidenced incapable of generating detailed and high-resolution views.} 
The aforementioned observations inspire us to consider a two-step pipeline: action2motion generates diverse \& natural 3D human motions from prescribed action categories, and motion2video proceeds to extract human character out of an additional input image, to rig, animate, and render to form 2D videos, as illustrated in Fig.~\ref{fig:teaser}.

In action2motion, we aim at generating diverse motions to traverse the motion space, and to cover various styles of individuals performing the same type of actions; meanwhile, each motion is expected to be visually plausible. This leads to our temporal variational auto-encoder (VAE) approach using Lie algebra pose representation. 
Inspired by the work of~\cite{denton2018stochastic} in generic video generation, here we leverage the posterior distribution learned from previous poses as a learned prior to gauge the generation of present pose; by tapping into the recurrent neural net (RNN) implementation, this learned prior also encapsulates temporal dependencies across consecutive poses. 
For pose representation, human pose could be characterized as a kinematic tree based on human body kinematics. There are multiple advantages of using Lie algebraic representation over the popular joint-coordinate representation: (i) Lie representation disentangles the skeleton anatomy, temporal dynamics, and scale information; (ii) it faithfully encodes the anatomical constraints of skeletons by following the forward kinematics~\citep{murray1994mathematical}; (iii) 
the dimension of Lie algebraic space corresponds exactly to the degree of freedom (DoF), which is more compact compared to joint-coordinate representation. In practice, the adoption of Lie representation notably mitigates the change-of-length and trembling phenomenons prevailing in joint coordinates representations; it also facilitates the generation of natural, lifelike motions, and simplifies the training process. Furthermore, a global and local movement integration module is used to infer the global pose trajectory from temporal articulations of body parts. This promotes consistence between local shape deformations and global motion trajectory (i.e. direction and velocity), especially when synthesizing locomotion actions such as walking and jumping. 

It is followed in our pipeline by motion2video, where a 3D character is extracted, rigged, animated according with stochastically generated motions, and rendered to form 2D videos.
In fact, animating 3D characters remains an open problem. A common strategy is to extract their 3D shapes and textures from a single input image. 
Prior efforts such as~\cite{weng2019photo} align the silhouette and texture of single image to a 3D human shape (e.g. SMPL~\citep{loper2015smpl}). Due to single input view, nonetheless, they fail to synthesize body textures of unseen views. \revision{Recent deep learning methods~\citep{lazova2019360,saito2019pifu,saito2020pifuhd,huang2020arch,zheng2021pamir} shed lights on reliable recovery of 3D surfaces and textures from single images. Meanwhile their results suffer from either low-fidelity, with input image resolution limited to at most 512$\times$512~\citep{saito2019pifu,huang2020arch,zheng2021pamir}}, or ill-posed texturing on occluded areas and novel view~\citep{saito2020pifuhd}. A simple strategy is developed in our work, leading to improved texture mapping in these cases.  

In summary, our main contributions are three-fold: 
first, a novel two-step pipeline of action2motion \& motion2video is proposed to address the challenging problem of 3D human motion \& video generation from action type and single image; 
second, a dedicated Lie Algebra based VAE framework is developed, capable of producing diverse life-like human motions from prescribed action categories; 
third, as part of our pipeline, an improved strategy is used in extracting 3D shapes and textures from single images, that is capable of synthesizing visually-appealing texture of unseen views. 
Moreover, an in-house 3D human motion dataset, HumanAct12, has been curated.

This paper differs from our preceding effort~\citep{guo2020action2motion} in a number of aspects: 
\begin{itemize}
    \item A more general problem of 3D human video generation is considered here, where the task of action2motion examined in~\cite{guo2020action2motion} becomes the first step of our solution pipeline. The motion2video step is entirely new from~\cite{guo2020action2motion}.
    \item A new local-global movement integration module is proposed, which significantly improves the synthesized 3D locomotion results when comparing to~\cite{guo2020action2motion}.
    \item A much broader and more thorough discussion is provided comparing to our short version~\citep{guo2020action2motion}. It also includes applications to latent interpolation, action transition, \revision{outpainting},  as well as evaluation of the synthesized motions from coarse- vs. fine-grained action categories.
\end{itemize}

\section{Related Work}\label{related work}

Our focus is to review literature related to generating video of human full-body motions, instead of the more generic theme of video generation~\citep{tulyakov2018mocogan,denton2017unsupervised,vondrick2016generating}. 
Our tally includes the discussion of action video generation (Sec.~\ref{rw:action2video}), the generation of human motions (Sec.~\ref{rw:action2motion}), motion transfer and rigid body animation (Sec.~\ref{rw:motion2video}). We also review related activities of VAE sequence modeling (Sec.~\ref{ss:vaeSM}), skeletal human pose representation (Sec.~\ref{ss:skeletonRepresentation}), and 3D human motion datasets (Sec.~\ref{ss:HumanMotionDatasets}).

\subsection{Action Video Generation}
\label{rw:action2video}
The task of generating human action videos has drawn research attentions very recently. 
In the work of~\cite{cai2018deep}, 2D human motions are generated from known actions, they are then synthesized into 2D videos frame-by-frame with U-Net~\citep{ronneberger2015u} and a dedicated image discriminator. In~\cite{yang2018pose}, based on an initial 2D pose extracted from a given image, a deterministic sequence of future 2D poses is produced for given action category; this pose sequence are subsequently used to guide video generation via adversarial training. A similar method is considered in~\cite{kim2019unsupervised}, where future 2D poses are instead generated stochastically with variational auto-encoder. 
These efforts focus on tiny pixel-wise video generation, and human poses are manipulated in 2D image space. 
A recent work~\citep{de2019generating} propose to generate 3D human videos directly from 3D game engine using scene composition rules and procedural animation techniques. Our work differs from this work in two folds: 1) \cite{de2019generating} generate 3D motions by extracting atomic motions from existing motion capture (MoCap) datasets, then stitches these atomic motions into action sequences through predefined rules. For example, a \textit{walking} animation involves repetitions of swinging a left leg, then swinging a right leg, as well as corresponding pendular arm movements. However, this process is fairly labor-intensive. In our work, diverse 3D actions are automatically produced from a learned generative model end-to-end; 2) \cite{de2019generating} animate artist-designed 3D avatars (rigid and clothed), while our method generates videos by rigging and animating characters with their 3D shapes and textures extracted from single 2D images.

\subsection{Human Motion Generation}
\label{rw:action2motion}
In addition to video generation, there are also research efforts focusing on synthesizing human motions, usually in the form of 2D or 3D skeletons, where the input could be of various forms, including but not limited to audio and text. 
One trendy research direction aims to generate deterministic motion sequences, which is typically realized by RNN models. 
For example, \cite{tang2018dance} and \cite{shlizerman2018audio} adopt LSTM models to translate music beats to body motion dynamics. In the efforts of~\cite{lin20181},~\cite{ahn2018text2action},~\cite{plappert2018learning}, and~\cite{yamada2018paired}, human motions are generated from textual descriptions through a encoder-decoder RNN model. ~\cite{ahuja2019language2pose} considers a closely related task of constructing a joint embedding space between sentences and human pose sequences. The work of~\cite{stoll2020text2sign} engages neural machine translation model with attention mechanism for text-to-sign-pose prediction. Similarly, a recurrent architecture is used in~\cite{pavllo2019modeling} to unfold an input global trajectory to locomotive humanoid movements.

To enable the stochasticity of human dynamics, deep generative models are also considered. \cite{habibie2017recurrent} propose a recurrent variational autoencoder model for global trajectory based locomotion generation. \cite{lee2019dancing} use GANs model to generate diverse movements from music signals. \cite{huang2020dance} explore a curriculum training strategy to allow variable sequence lengths. In \cite{cai2018deep}, a two-stage GAN framework is proposed to generate 2D human motion progressively. To synthesize human motions from scratch, \cite{zhao2020bayesian} and \cite{zhao2018adversarial} make use of Bayesian inference; the work of \cite{yan2019convolutional} instead considers a combined strategy of graph convolutional networks and GANs. The recent work of~\cite{xu2020hierarchical} synthesizes novel motions by free combination of style and content codes extracted from existing MoCap library. 


\subsection{Motion Transfer and Rigid Body Animation}
\label{rw:motion2video}

Motion transfer is a traditional topic, aiming to transfer human motions from a source object to target. Recent deep learning based efforts typically consider 2D pixel-wise approaches, where mappings from source and target are based on local pixels or 2D patches. 
\cite{wang2018video} and \cite{chan2019everybody}, for example, directly learn to map between human poses and appearances of one specific source subject. 
The aim of~\citep{siarohin2019animating,wang2019few,lee2019metapix,liu2019liquid} is to work toward a more general problem of driving an arbitrary target image with a source 2D pose sequence or videos. This is often realized by establishing connections between the source pose sequence and the target textured shape extracted from an given image, followed by warping the reference image to form the target video frame-by-frame. Although assembling promising results, the mainstream pixel-wise approaches nonetheless possess a number of limitations, including its innate difficulties in dealing with changing views or lifting to 3D motion spaces, as well as the level of complications in producing high-resolution and sharp images. \revision{The works of~\citep{villegas2018neural,aberman2020skeleton} also consider a similar task, where motions from the source 3D character are re-targeted to 3D characters with different skeletons (e.g. joint number, bone lengths). Meanwhile, the 3D shapes of these target characters have been artistically designed and well-rigged ahead of time.}

Meanwhile, it has also been a continuous line of research on rigid body animation of 2D/3D human characters that is especially empowered by advances in computer graphics techniques. 
Early work such as~\cite{zhou2012image} uses a simple pose-retrieval framework, where a segmented garment database indexed by 2D skeleton poses is built for online searching during human image animation. 
Rigged human models are exploited in later endeavors for articulated object animation. In~\cite{hornung2007character}, characters extracted from 2D pictures are driven as-rigid-as-possible by external 3D MoCap sequences. At intermediate steps, a 2D mesh with 2D skeleton is constructed for the shape extracted from input image. \cite{weng2019photo} further lifts this animation process into 3D space. Specifically, a semi-naked SMPL template is drawn out of 2D images, and deformed to a rigged 3D mesh model with boundary that closely matches to the human silhouette in input image. \revision{The recent work of~\cite{huang2020arch} learns to directly predict a 3D animatable clothed human shape from a single image.} 

\subsection{VAE in Sequence Modeling}\label{ss:vaeSM}
Variational autoencoder~\citep{kingma2013auto} are the encoder-decoder neural nets trained by maximizing the marginal data likelihood with variational methods. It has been widely used in the so-called deep generative models as a powerful learning technique in addressing various learning scenarios, including conditional generation~\citep{sohn2015learning}, semi-supervised learning~\citep{kingma2014semi,siddharth2017learning}, controllable generation~\citep{cheng2020segvae}, few-shot learning~\citep{schonfeld2019generalized}, disentangle representation learning~\citep{ding2020guided,zhu2020s3vae,higgins2016beta} and VAE-GAN architecture~\citep{larsen2016autoencoding}.

To work with sequential data, VAEs are typically plugged in a recurrent network model, e.g. GRU and LSTM. Variational RNN~\citep{chung2015recurrent}, a pioneer work, uses vanilla RNN to model temporal dependencies in intermediate time-frames. The RNN output of previous frame is used in generating posterior and prior distributions, as well as the follow-up decoding process. Variational RNN has been particularly favored in speech generation and handwriting character generation. 
\cite{bowman2015generating} and \cite{yang2017improved} investigate the LSTM-based VAE for NLP modelling based on a sequence-to-sequence architecture, where the sequence encoder predicts a posterior distribution, from which the sequence decoder samples a latent vector and reconstruct the sequence. 
%
More specifically, temporal VAE models has been considered in motion and video generation. \cite{marwah2017attentive} consider generating videos from textual caption, which is incorporated as semantic attentive vectors and fed to their temporal VAE. In VideoVAE~\citep{he2018probabilistic}, on the other hand, a structured latent unit is devised to model conditional factors including motion category and an initial frame to complete the rest frames. To predict future frames under uncertainty, \cite{denton2018stochastic} inspect the use of two separate RNNs to capture temporal dependencies of conditional posterior and prior spaces. 
Similar network structure is also scrutinized in~\cite{wang2019point}, where it is extended to synthesize videos with pre-specified start and end frames. 
%
In terms of 3D motion prediction, given a start human pose, \cite{habibie2017recurrent} complete the rest 3D human motion with a LSTM-based VAE model. 
In \cite{yan2018mt}, similar model is engaged to learn the transition from observed sequence to future sequence for stochastic motion forecasting. A very recent work by~\cite{aliakbarian2020stochastic} adopts VAE and a mix-and-perturbation strategy to statistically predict future motions.

\subsection{Skeletal Human Pose Representation}\label{ss:skeletonRepresentation}
A number of human pose representations have been considered over the years. The most-often used option is the joint-coordinate representation~\citep{han2017space,hussein2013human} that directly characterizes the human pose by an ordered sequence of 2D/3D joint coordinates. It has a few variants: \cite{wang2012mining} consider incorporating the pair-wise relative positions of neighboring joints; meanwhile, only those informative joints are utilized in~\cite{chaaraoui2014evolutionary}. Part-based method is another line of pose representation. Specifically, a human pose is modeled as a ordered list of body parts. For example, in~\cite{yacoob1999parameterized}, human body is divided into five main parts (i.e. torso and four limbs); pose sequences are then formulated by the displacement and rotations of body parts over time. Alternatively, the work of \cite{muller2007information} models the temporal information using dynamic time warping. Finally, Lie group or axis-angle based representation~\citep{gavrila1995towards,vemulapalli2014human,huang2017deep,xu2017lie,liu2019towards,pavllo2019modeling} characterizes the skeleton as a kinematic tree, with its articulations realized by forward kinematics. 

\subsection{3D Human Motion Datasets}\label{ss:HumanMotionDatasets}
CMU MoCap~\citep{cmu2003mocap} and HDM05~\citep{muller2007mocap} have more than 100,000 3D poses and 2,000 3D motion sequences that are associated with succinct textual descriptions. Unfortunately, the motions are markedly uneven-distributed over action categories. UTKinect-Action~\citep{xia2012view} and MSR-Action3D~\citep{li2010action}, on the other hand, have much smaller tally of motion sequences. NTU-RGBD~\citep{liu2019ntu} is by far the largest human motion dataset, consisting of over 100,000 motions belonging to 120 classes. Nevertheless, the joint positions acquired from Microsoft Kinect-I cameras are notably inaccurate. These observations motivate us instead curating our in-house 3D human action dataset, \DN, as well as revamping the pose annotations of NTU-RGBD.


\section{Preliminary Backgrounds}

\subsection{Variational Auto-Encoder}

Variational auto-encoder(VAE)~\citep{kingma2013auto} consists of an encoder and a decoder, which are normally two separate neural networks. Its goal is to learn a $\theta$-parameterized generative model, $p_{\theta}(\mathbf{x}, \mathbf{z})$, over data $\mathbf{x}$ and latent variables $\mathbf{z}$. Technically, the learning objective is to maximize the likelihood function of $\mathbf{x}$, which could be further formulated as a marginal likelihood with regard to the latent variable $\mathbf{z}$, $p_\theta(x)=\int_\mathbf{z} p_\theta(\mathbf{x}|\mathbf{z})p_\theta(\mathbf{z})$. Following the variational principle, a $\phi$-parameterized neural network(i.e. encoder), $q_\phi(\mathbf{z}|\mathbf{x})$, is engaged to approximate the unknown posterior distribution $p_\theta(\mathbf{z}|\mathbf{x})$. We thus obtain the the following evidence lower bound (ELBO) to our data likelihood function:
\begin{equation}
    \begin{aligned}
    \log  p_\theta(\mathbf{x}) &= \log \int_\mathbf{z} p_\theta(\mathbf{x}|\mathbf{z})p(\mathbf{z}) \\
            &\geq \mathbb{E}_{q_\phi(\mathbf{z}|\mathbf{x})}\log p_\theta(\mathbf{x}|\mathbf{z}) \\
            & \, \, \, \, \, \,- D_{\mathrm{KL}}(q_\phi(\mathbf{z}|\mathbf{x})\parallel p(\mathbf{z})). 
    \end{aligned}
\end{equation}
The first ELBO term encourages the generated samples to be sufficiently close to the real samples; the second term penalizes KL-divergence between the prior and the approximated posterior distribution. Subsequently, the original objective of maximizing the data likelihood over data $\mathbf{x}$ becomes that of maximizing over the $\theta$- and $\phi$-parameterized ELBO function.
In \cite{sohn2015learning}, a follow-up \textit{conditional} variational auto-encoder (CVAE) framework is conceived by introducing a conditional variable, $\mathbf{y}$, as
\begin{equation}
    \begin{aligned}
    \log  p_\theta(\mathbf{x}|\mathbf{y}) &= \log \int_\mathbf{z} p_\theta(\mathbf{x}|\mathbf{z,y})p(\mathbf{z|y}) \\
            &\geq \mathbb{E}_{q_\phi(\mathbf{z}|\mathbf{x,y})}\log p_\theta(\mathbf{x}|\mathbf{z,y}) \\
            & \,\,\,\,\,\,- D_{\mathrm{KL}}(q_\phi(\mathbf{z}|\mathbf{x,y})\parallel p(\mathbf{z})). 
    \end{aligned}
\end{equation}

\subsection{Lie Groups and Lie Algebras}

In what follows, we provide a succinct introduction of Lie groups and Lie algebra basics. Interested readers may refer to~\citep{murray1994mathematical} for more details.

\textbf{Lie groups.} Mathematically, a Lie group is a group as well as a smooth manifold. 
3D rotation transformations, also known as the Special Orthogonal group, $\mathrm{SO3} = \{R\in \mathbb{R}^{3\times 3}|R^\top R=I, \mathrm{det}(R)=+1\}$, is a classical example of Lie group. Moreover, the product of multiple $\mathrm{SO3}$ groups (i.e. a kinematic chain) is still a Lie group. In other words, for a tree-structured human skeleton model, each of the kinematic chains corresponds to a point in Lie group $\mathrm{SO}(3)\times \mathrm{SO}(3)\times \cdots \times \mathrm{SO}(3)$. As a consequence, it is usually far from being trivial in terms of optimization in such a curved space. We instead work in its tangent space, also known as Lie algebra $\mathfrak{so}(3)$-- being a flat space, our familiar linear algebra techniques could work again.

\textbf{Lie algebras.} The tangent space of Lie group $\mathrm{SO(3)}$ at identity $\mathrm{I}_3$ is referred to as its Lie algebra $\mathfrak{so}(3)$. Each element of $\mathfrak{so}(3)$ is in the form of a $3\times3$ skew-symmetric matrix $\hat{W}$, as
\begin{equation}
    \hat{W} = 
    \left(
    \begin{array}{ccc}
        0 & -w_3 & w_2  \\
        w_3 & 0 & -w_1  \\
        -w_2 & w_1 & 0
    \end{array}
    \right),
\end{equation}
which essentially spans a 3-dimensional vector space, $\mathbf{w}=(w_1, w_2,$ $ w_3)^\top\in \mathbb{R}^3$. 

\textbf{Exponential map.} To map from a Lie algebra element $\hat{W}\in\mathfrak{so}(3)$ to a point in the manifold (i.e. Lie group), $R\in \mathrm{SO}(3)$, an exponential map $\exp: \mathfrak{so}(3) \rightarrow \mathrm{SO}(3)$ is formulated as
\begin{equation}
    R = \exp{(\hat{W})} = \mathrm{I} + \frac{\sin(\|\mathbf{w}\|)}{\|\mathbf{w}\|}\hat{W} + \frac{1-\cos(\|\mathbf{w}\|)}{\|\mathbf{w}\|^2}\hat{W}^2.
\end{equation}
Here $\|\cdot\|$ is a vector norm. Since $\mathbf{w}$ is periodically mapped to $R$, in practice we normally limit $\mathbf{w}$ by its norm within the range of $[-\pi, \pi]$. Its inverse map, the logarithm map $\log (\mathrm{SO}(3))$: $ \mathrm{SO}(3) \rightarrow \mathfrak{so}(3)$ map be similarly constructed. 

\section{Our Approach}
The pipeline of our approach, \textbf{action2video}, consists of two steps: step one (action2motion) synthesizes human pose sequences from a prescribed action category (Sec.~\ref{action2motion}); step two (motion2video) extracts a specific 3D human shape and texture from a reference image to render the generated motions into 2D videos (Sec.~\ref{motion2video}).

\subsection{Step One: Action2Motion}
\label{action2motion}
Our action2motion framework comprises a temporal VAE (Sec.~\ref{temporal vae}) with a Lie algebra based representation (Sec.~\ref{Lie algebra}). We also investigate four strategies to decode neural hidden unit to obtain global 3D positions of motions (Sec.~\ref{archit} and Sec.~\ref{pose decoding}). 


\subsubsection{Disentangled Representation with Lie Algebra}
\label{Lie algebra}

\begin{figure}[t]
  \centering
  \includegraphics[width=0.7\linewidth]{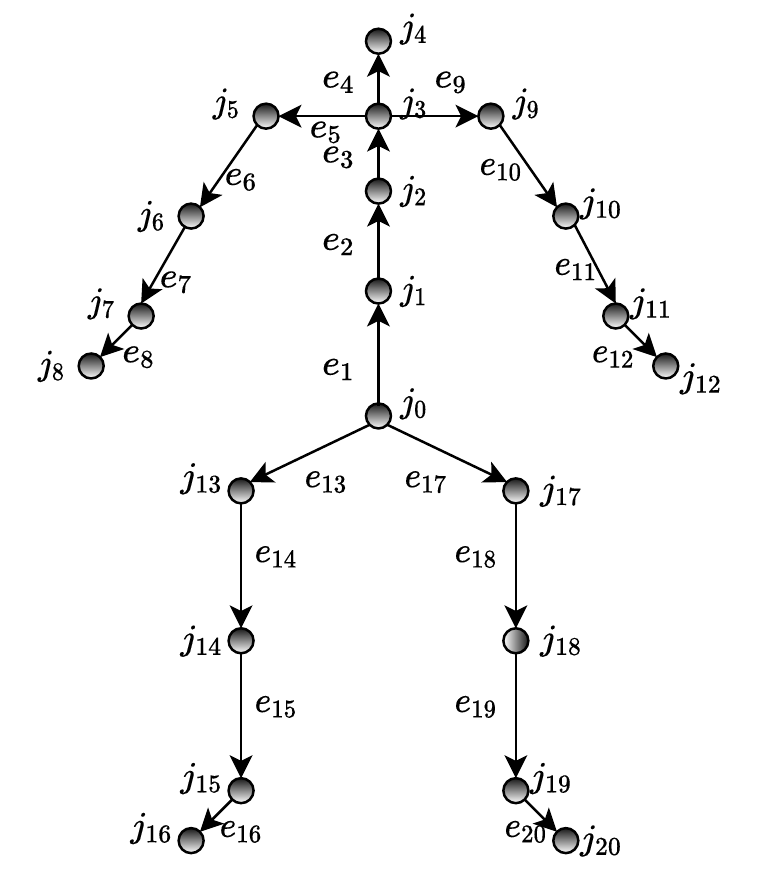}
  \caption{An example of human skeleton which consists of 21 joints and 20 body parts. }
  \label{fig:skeleton}
  \vspace{-0.3cm}
\end{figure}

As shown in Fig.~\ref{fig:skeleton}, a human pose could be characterized in the form of a kinematic tree that consists of five kinematic chains: main spine and four limbs. 
Meanwhile, this skeleton model is formed by $N$ oriented edges (i.e. bones) $E=\{e_1,\ldots,e_N\}$ that interconnect $N+1$ joints. By incorporating Lie algebraic apparatus, motion of 3D joints could be decomposed into three parts: skeleton anatomical information, motion trajectories, and bone lengths.

\label{method}
For each skeletal bone, $e_n$, a local coordinate is attached, with the bone itself being aligned with the x-axis and its starting joint being stuck to the coordinate origin. The relative 3D locations between two consecutive bones could be modeled as a series of 3D rigid transformations. Specifically, given two connected bones $e_n$ and $e_{n+1}$ along a kinematic chain, a joint $\mathbf{c}=(x, y, z)^\top$ in the local coordinate of $e_n$ amounts to a transformed location $\mathbf{c'}=(x', y',z')^\top$ in the local coordinate of $e_{n+1}$, by exercising the following transformation
\begin{equation}
    \left( \begin{array}{c}
         \mathbf{c'}  \\
         1 
    \end{array} \right)= \left(
    \begin{array}{cc}
        \mathbf{R}_{n} & \mathbf{d}_{n}  \\
        0 & 1
    \end{array}\right)\left(\begin{array}{c}
         \mathbf{c}  \\
          1
    \end{array}\right).
\end{equation}
Here, $\mathbf{R}_n \in \mathbb{R}^{3\times3}$ is a rotation matrix, $\mathbf{d}_n = (b_n, 0, 0)^\top$ $ \in \mathbb{R}^3$ a translation vector along x-axis, and $b_n$ the length of bone $e_n$. 

For a 3D rotation matrix $R \in \mathrm{SO(3)}$, the associated Lie algebraic vector $\mathbf{w} \in \mathfrak{so}(3)$ is an axis-angle vector. For a human skeleton, the exact degree of freedom (DoF) of a axis-angle vector is determined by the rotation orientations of two successive bones, and is up to 3. For example, if two bones are oriented in the same or reverse direction, $\mathbf{w}$ is a zero vector with 0 DoF; if one bone only rotates along one axis, then the DoF reduces to 1.

\textbf{Mapping Lie algebra parameters to 3D positions.} Now we focus on an articulate object with $K$ kinematic chains; assume the $k$-th chain have $m_k$ joints, with each joint parameterized by a 3-dimensional $\mathfrak{so}(3)$ vector, $\mathbf{w}_i^k, i\in\{1,2,\ldots,m_k\}$. A human pose is thus represented by composition of Lie algebra vectors of joints/bones on kinematics chains, $\mathbf{p}_{\mathrm{Lie}} = ({w_1^1}^\top, \ldots, {w_{m_1}^1}^\top,$ $ \ldots,{w_1^K}^\top, \ldots,$ $ {w_{m_K}^K}^\top)$. Now, the 3D position of a joint $i$ in a chain $k$, $\mathbf{J}_i^k$, is  obtained following a exponential map of the Lie algebraic values, also known as \textit{forward kinematics}, as 
\begin{equation}
    \mathbf{J}_i^k = \left[\prod_{j=0}^{i-1}\exp(\hat{W}_j^k)\right]\mathbf{d}_i^k + \mathbf{J}_{i-1}^k.
\end{equation}
Here $\mathbf{d}_i^k=(b_i^k, 0, 0)$, with $b_i^k$ representing the bone length of $e_i^k$. In addition, forward kinematics typically starts from a root joint whose position $\mathbf{J}_0\in \mathbb{R}^3$, and Lie algebraic values $\hat{W}_0$ stand for the global location and orientation of the entire human body. In our representation, the global location $\mathbf{J}_0$ is independent from the pose. 
Therefore, given a motion with $T$ successive poses, the sequence $(\mathbf{J}_{0,1}, \ldots, \mathbf{J}_{0, T})\in \mathbb{R}^{3\times T}$ makes up the body motion trajectory, with $\mathbf{J}_{0, t}$ denoting its global location at frame $t$.

\label{forward kinematics}

Accordingly, the 3D coordinates vector of a body pose, formally denoted as $\mathbf{p} = ({\mathbf{J_1}^1}^\top, \ldots, $ ${\mathbf{J_{m_1}}^1}^\top, \ldots $ $, {\mathbf{J_1}^K}^\top, \ldots, {\mathbf{J_{m_K}}^K}^\top)$ could be obtained by the joint-wise forward kinematics of a composition of \textit{bone lengths}, \textit{root position}, and \textit{Lie algebraic vector}. For simplicity, we denote this mapping as $\mathbf{\Gamma}(\mathbf{p}_\mathrm{Lie}): \mathbf{p}_\mathrm{Lie} \rightarrow \mathbf{p}$.
Overall, 
a human \textit{motion} is represented by three parts:
\begin{itemize}
    \item Lie algebra parameters $\mathbf{M}_\mathrm{Lie}=\left(\mathbf{p}_\mathrm{Lie}^1, \ldots,\mathbf{p}_\mathrm{Lie}^T \right)$. 
    \item Root trajectory $(\mathbf{J}_{0,1}, \ldots, \mathbf{J}_{0, T})$: root trajectory could be represented by either absolute root locations or relative translations between consecutive root locations. The latter works better in our setting.
    \item Bone lengths $(b_0, \ldots, b_N)$: due to the invariant nature of bone lengths of human skeleton, the skeleton bone lengths are acquired from typical real-life human bodies, and are fixed over time. This also reciprocally enables us to generate motions with controllable body scales by manipulating the bone lengths.
\end{itemize}

\subsubsection{Conditioned Temporal VAE}

\begin{figure*}[t]
  \centering
  \includegraphics[width=0.8\linewidth]{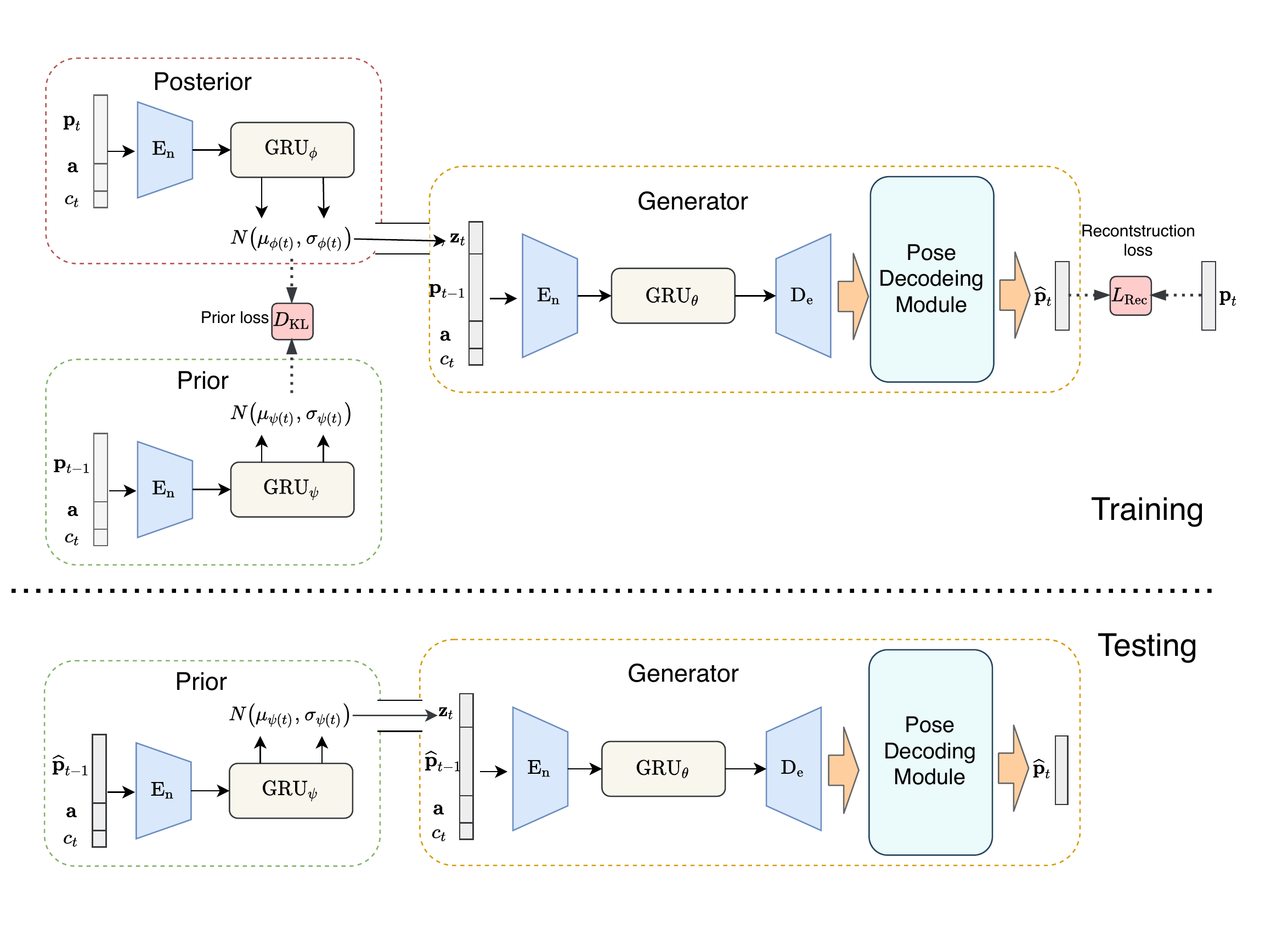}
  \caption{Visual diagram of action2motion, the first step in our pipeline. Top row shows the training phase: at time $t$, the posterior and prior networks take as input a concatenation of three parts - action category $a$, time counter $c_t$ and immediate pose vector ($p_t$ or $p_{t-1}$). The generator receives an addition latent vector $\mathbf{z}_t$ that is sampled from the learned posterior distribution. Afterwards, the 3D joints of current pose is obtained from the decoder of generator through \textit{pose decoding module}. Bottom row depicts the testing phase: a latent vector is alternatively sampled from the prior distribution, which triggers the aforementioned process in generating 3D pose sequences.}
  \label{fig:architecture}
\end{figure*}

\begin{figure*}[t]
  \centering
  \includegraphics[width=\linewidth]{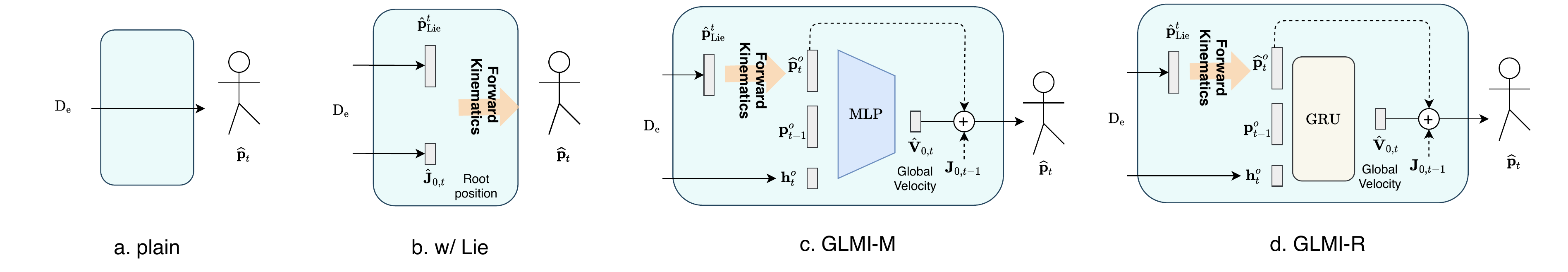}
  \caption{Four variants of the pose decoding module conceived in our work: (a) direct generation of 3D joint positions; (b) generation with Lie algebraic representation; and (c)-(d) \textit{global and local movement integration} (\textbf{GLMI})-based generation with Lie algebraic representation, implemented by multi-layer perceptron (GLMI-M) or GRU (GLMI-R).}
  \label{fig:pose_decoding}
\end{figure*}

\label{temporal vae}

Consider a real motion or pose sequence $\mathbf{M}=\left(\mathbf{p}_1, \ldots,\mathbf{p}_T\right)$. Our temporal VAE aims to maximize the likelihood of the pose sequence $\mathbf{M}$. 
At time $t$, a posterior network $q_\phi(\mathbf{z}_t|\mathbf{p}_{1:t})$ approximates the true posterior distribution conditioned on $\mathbf{p}_{1:t-1}$. Then, with sampled latent variables $\mathbf{z}_{1:t}$ and previous states $\mathbf{p}_{1:t-1}$, our RNN generator $p_\theta(\mathbf{p}_t|\mathbf{p}_{1:t-1}, \mathbf{z}_{1:t})$ reconstructs the current pose $\mathbf{p}_t$. 
This leads to the following variation lower bound:

\begin{equation}
    \begin{aligned}
    \log  p_\theta(\mathbf{M}) &\geq \sum_t \bigg[\mathbb{E}_{q_\phi(\mathbf{z}_t|\mathbf{p}_{1:t})}\log p_\theta(\mathbf{p}_t|\mathbf{p}_{1:t-1}, \mathbf{z}_{1:t}) \\ 
            &\,\,\,\,\,-D_{\mathrm{KL}}\left(q_\phi(\mathbf{z}_t|\mathbf{p}_{1:t})\parallel p(\mathbf{z}_t)\right)\bigg].
    \end{aligned}
\end{equation}
Note at time $t$, our RNN module takes as input the immediate past frame $\mathbf{p}_{t-1}$ and  $\mathbf{z}_t$. The influence from previous time slices $\mathbf{p}_{1:t-2}$ and $\mathbf{z}_{1:t-1}$ lies in the ability of RNN module capturing long-term temporal dependencies.

In terms of the prior $p(\mathbf{z}_t)$, one option is to consider an identity Normal distribution, $\mathcal{N}(0, \mathbf{I})$. 
This is unsuitable though for the motion generation problem, as the pose variation varies over time. 
Take \textit{running} motions as example, the temporal pose variances are typically relatively small, which however could become significantly larger when e.g. the runner makes a U-turn. 
Inspired by the observation that the variation of present pose is highly correlated to its past time-steps~\citep{denton2018stochastic}, we model its prior by a neural network that conditions on its previous steps $\mathbf{p}_{1:t-1}$, $p_\psi(\mathbf{z}_t|\mathbf{p}_{1:t-1})$. This leads to a re-formulation of the ELBO objective function
\begin{equation}
    \begin{aligned}
    \log  p_\theta(\mathbf{M}) &\geq \sum_t \bigg[\mathbb{E}_{q_\phi(\mathbf{z}_t|\mathbf{p}_{1:t})}\log p_\theta(\mathbf{p}_t|\mathbf{p}_{1:t-1}, \mathbf{z}_{1:t}) \\ 
            &\,\,\,\,\,-D_{\mathrm{KL}}\left(q_\phi(\mathbf{z}_t|\mathbf{p}_{1:t})\parallel p_\psi(\mathbf{z}_t|\mathbf{p}_{1:t-1})\right)\bigg],
    \end{aligned}
\end{equation}
where the distance penalty between prior and posterior distributions further encourages temporal consistency.

\subsubsection{Architecture of Action2Motion}
\label{archit}

Our action2motion step consists of three main components: posterior network, prior network, and generator, which are shown in Fig.~\ref{fig:architecture}. The input vector contains the following parts: the pose vector $\mathbf{p}_t$ or $\mathbf{p}_{t-1}$, an one-hot vector $\mathbf{a}$ to encode action category, and $c_t \in [0,1]$, a time-counter to keep record of where we are in the sequence generation progress. 
%
As depicted in Fig.~\ref{fig:architecture}, during training, a noise vector is sampled from the posterior distribution $q_\phi(\mathbf{z}_t|\cdot)$, and fed into the generator, which then produces the final 3D pose prediction by running through the pipeline of encoder $\mathrm{E_n}$, GRU unit $\mathrm{GRU}_\theta$, decoder $\mathrm{D_n}$, and \PDM. In testing, as the real data $\mathbf{p}_t$ is not available, $\mathbf{z}_t$ is instead sampled from the learned prior distribution, $p_\psi(\mathbf{z}_t|\cdot)$.

Specifically, our encoder $\mathrm{E_n}$ and decoder $\mathrm{D_n}$ are composed of linear fully connected layers with different weights, and updated with the whole network. Moreover, our posterior network ($q_\phi$) and prior network ($p_\psi$) utilize the same architecture, but with different parameters. They are respectively described as: 
\begin{equation}
    \begin{aligned}
    \mathbf{h}_t &= \mathrm{E_n}(\mathbf{p}_t, \mathbf{a}, c_t),\,\,\, c_t = \frac{t}{T} \\
    \left( \mu_\phi(t), \sigma_\phi(t) \right) &= \mathrm{GRU}_\phi(\mathbf{h}_t)\\
    \end{aligned}
\end{equation}
and
\begin{equation}
    \begin{aligned}
    \mathbf{h}_{t-1}' &= \mathrm{E_n}(\mathbf{p}_{t-1}, \mathbf{a}, c_t),\,\,\, c_t = \frac{t}{T} \\
    \left( \mu_\psi(t), \sigma_\psi(t) \right) &= \mathrm{GRU}_\psi(\mathbf{h}_{t-1}'). \\
    \end{aligned}
\end{equation}
Further investigation of the pose decoding module is provided in the following section.

\subsubsection{Pose Decoding}
\label{pose decoding}
Fig.~\ref{fig:pose_decoding} illustrates the four pose decoding variants investigated in our work. The most straightforward and commonly-used approach is Fig.~\ref{fig:pose_decoding}(a), where the 3D joint locations are directly and simultaneously regressed from the decoder. It however contains redundant parameters, and does not follow the kinematics law that dictates the 3D articulations of the body skeleton. 
Alternatively, the Fig.~\ref{fig:pose_decoding}(b) variant incorporates Lie algebraic representation, which is the one adopted in our previous work~\citep{guo2020action2motion}. 
The decoder here contains two vectors, skeletal Lie algebraic values $\hat{\mathbf{p}}_\mathrm{Lie}^t$, and global root position $\hat{\mathbf{J}}_{0,t}$. 
The final 3D joints are produced by \textit{forward kinematics} (see Sec.~\ref{forward kinematics}). Though working well for many motion scenarios, it encounters issues when local body movements and global motions are highly correlated. Take action \textit{walk} for example, the instantaneous velocity of walking is significantly affected by the movement of \textit{legs}; independently generating global and local body motions is observed to lead to e.g. sliding-feet phenomenon, as depicted in Fig.~\ref{fig:locomotion}.

\textbf{Global and local movement integration.} 
Existing efforts in motion forecasting or generation usually predict \textit{only} relative body joint positions, this is, relative to the root joint, at the cost of neglecting the global motion all together~\citep{wang2020learning,yan2019convolutional,liu2019towards,xu2017lie}. In other words, the root joint of human full-body is fixed to coordinate origin during the entire motion sequence. Recently, \cite{adeli2020socially} consider global motion by directly enforcing MSE or $\ell_2$ loss between predicted and ground-truth root joint locations, which is similar to the Fig.~\ref{fig:pose_decoding}(a) variant. 

Intuitively, the transition between two consecutive poses, measured by the displacement of the root joint in the two frames, is highly correlated to the body gesture of these two poses. Consider a person who is walking on a flat ground, his walking pace depends upon how wide his legs span. This inspires us to propose a \textit{global and local movement integration unit} (\textbf{GLMI}) which, rather than predicting global transition and local joints concurrently, will first generate relative poses, then infer global motion from consecutive local poses, as illustrated in Fig.~\ref{fig:pose_decoding}(c). Here $\hat{\mathbf{p}}_\mathrm{Lie}^t$ is the Lie parameter vector produced by the generator, which is then transformed to 3D joint locations $\hat{\mathbf{p}}_t^o$ through forward kinematics; $\mathbf{p}_{t-1}^o$ is the offset value of 3D coordinates of previous pose; $\mathbf{h}_t^o$ is a hidden vector containing upstream information. 
The three vectors are fed into a fully connected layer, MLP, which then produces the velocity (i.e. relative translation) $\hat{\mathbf{V}}_{0,t}$ at time $t$. Finally, the 3D global position $\hat{\mathbf{p}}_t$ could be obtained by summation of the three components: root position of previous pose $\mathbf{J}_{0, t-1}$, estimated velocity $\hat{\mathbf{V}}_{0,t}$, and the current local pose $\hat{\mathbf{p}}_t^o$. Mathematically, this process is expressed as
\begin{equation}
    \begin{aligned}
        \left(\mathbf{\hat{p}}_\mathrm{Lie}^t, \mathbf{h}_t^o \right) & = \mathrm{D_e}(\mathbf{h}_t^\theta)\\
        \mathbf{\hat{p}}_t^o &= \mathbf{\Gamma}(\mathbf{\hat{p}}_\mathrm{Lie}^t)\\
        \mathbf{\hat{V}}_{0,t} &= \mathrm{MLP}(\mathbf{\hat{p}}_t^o, \mathbf{p}_{t-1}^o, \mathbf{h}_t^o) \\
        \mathbf{\hat{p}}_t &= \mathbf{\hat{p}}_t^o + \mathbf{J}_{0, t-1} + \mathbf{\hat{V}}_{0, t}.
    \end{aligned}
\end{equation}

To further capture the temporal dependency of a global trajectory, another version of GLMI is also proposed, with the backbone of MLP replaced by recurrent units, GRU, as presented in  Fig.~\ref{fig:pose_decoding}(d). Besides, a trajectory alignment loss between the predicted velocities $\hat{\mathbf{V}}_{0,t}$ and real velocities $\mathbf{V}_{0,t}$ is also introduced, to 
encourage accurate velocity estimation.
Among these variants, the GLMI-M variant is found to produce the overall best results, and is utilized in our approach by default.

\begin{figure*}[t]
  \centering
  \includegraphics[width=\linewidth]{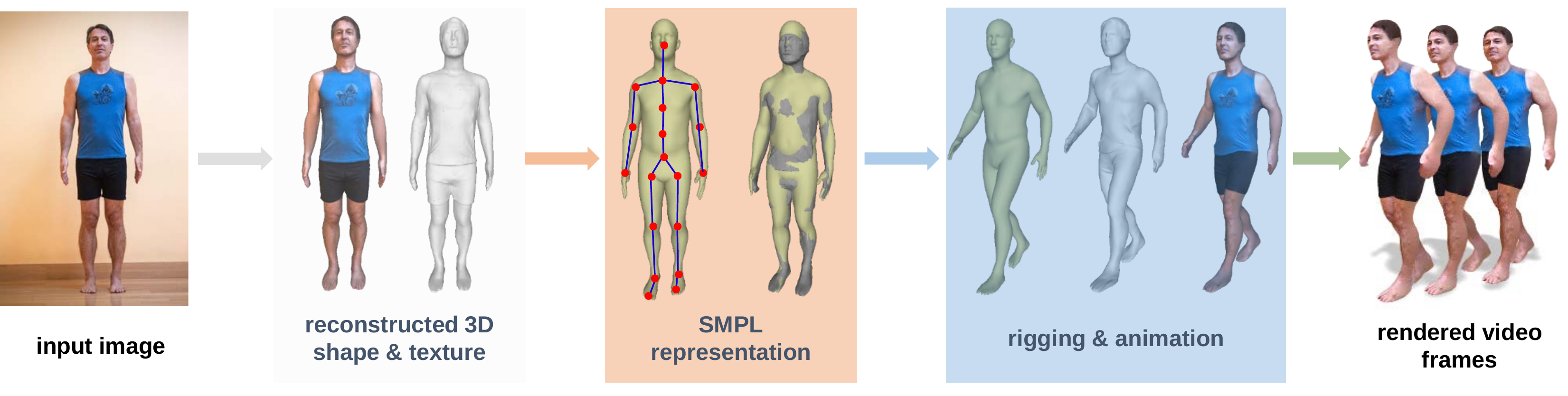}
  \caption{Illustration of the motion2video process. Shapes and textures of 3D human characters are extracted from single 2D images, that are rigged, animated with motions generated from the action2motion step, and rendered to produce final videos.
  }
  \label{fig:video_pipleline}
  \vspace{-0.3cm}
\end{figure*}
\subsubsection{Final Objective}

To summarize, our final objective function becomes

\begin{equation}
    \begin{aligned}
    \mathcal{L}_{\theta,\phi,\psi} &= -\sum_{t=1}^T \bigg[ \mathbb{E}_{q_\phi(\mathbf{z}_t|\mathbf{p}_{1:t}, \mathbf{a}, c_t)}\log p_\theta(\mathbf{p}_t|\mathbf{p}_{1:t-1}, \mathbf{z}_{1:t}, \mathbf{a}, c_t) \\ &\,\,\,\,\,-\lambda_{kl} D_{\mathrm{KL}} \left(q_\phi \left(\mathbf{z}_t|\mathbf{p}_{1:t}, \mathbf{a}, c_t \right)\parallel p_\psi \left(\mathbf{z}_t|\mathbf{z}_{1:t-1}, \mathbf{a}, c_t \right)\right)\\ &\,\,\,\,\,-\lambda_{align}\|\mathbf{V}_{0,t}-\hat{\mathbf{V}}_{0,t}\|_2 \bigg],
    \end{aligned}
    \label{eq:objective}
\end{equation}
where $\lambda_{kl}$ and $\lambda_{align}$ are two tuning parameters to trade-off among reconstruction error $\mathcal{L}_{rec}$, KL-divergence, and trajectory alignment loss. 
Empirically, a larger $\lambda_{kl}$ is observed to enhance the quality of generated motions but may decrease their diversity; and vice versa for a smaller $\lambda_{kl}$. 

For the reconstruction error (the first term in Eq.~\eqref{eq:objective}), the per-joint loss suggested in~\cite{aksan2019structured} is considered, as
\begin{equation}
    \mathcal{L}_{rec}(\mathbf{p}_t, \mathbf{\hat{p}}_t) = \sum_{k=1}^{N+1}\|\mathbf{J}_{k,t}-\hat{\mathbf{J}}_{k,t}\|_2.
\end{equation}
Here $N+1$ denotes the number of skeletal joints. 

In our work, the trajectory alignment loss is only used in the methods of Fig.~\ref{fig:pose_decoding}(c) and (d), 
where the models are trained with the re-parameterization trick of~\cite{kingma2013auto}.

\subsubsection{Training Strategy}
One common issue in sequence modeling is the discrepancy of information exposure during training vs. testing phases. For example, in a RNN model, a \textit{ground-truth} pose is taken as input to generate next pose in training; while in testing phase, a \textit{generated} pose is used instead to produce next pose. To mitigate the issue, a mixed training strategy is adopted here, that chooses whether to use (or not to use) \textit{teacher forcing}~\citep{bengio2015scheduled}  by randomly draws from a Bernoulli distribution, $V \sim \mathrm{Bernoulli}(p_{\mathrm{tf}})$. In particular, teacher forcing is chosen for the entire sequence $\mathbf{p}_{1:T}$ if $V$ is 1, and not if otherwise.
As a boundary condition in generating the initial pose $\hat{\mathbf{p}}_1$, its previous pose input $\mathbf{p}_0$ for the prior network ($q_\psi$) is a zero vector. In addition, \textit{curriculum learning}~\citep{bengio2009curriculum} is used in the training phase that is to progressively increase the value of $\lambda_{kl}$.

\subsection{Step Two: Motion2Video}\label{motion2video}

Recall in step one of our approach, action2motion, diverse motions are generated from prescribed action categories. At this point, a motion is shown as a sequence of 3D skeletal articulations. To produce videos, it remains to settle the full-body shapes and textures of the involved human characters. This is addressed in step two, motion2video, where a specific setup is conceived: a reference person image is presented as input, from which 3D shape and texture of the person are extracted; this is followed by rigging and animating the characters with synthesized motions from the action2motion step, and rendering to generate final 2D videos.  
Unlike existing motion transfer methods~\citep{chan2019everybody,liu2019liquid,wang2019few} that emphasize in 2D space, our work advocates a fully 3D approach, and we claim our 3D-enabled modelling choice helps to preserve the geometric and appearance aspects in the final video production. Fig.~\ref{fig:video_pipleline} illustrates the components in our motion2video process that is to be detailed in the following subsections. 

\subsubsection{Human Shape Reconstruction from a Single 2D Image} 
\begin{figure*}[tb]
	\centering
	\includegraphics[width=0.9\linewidth]{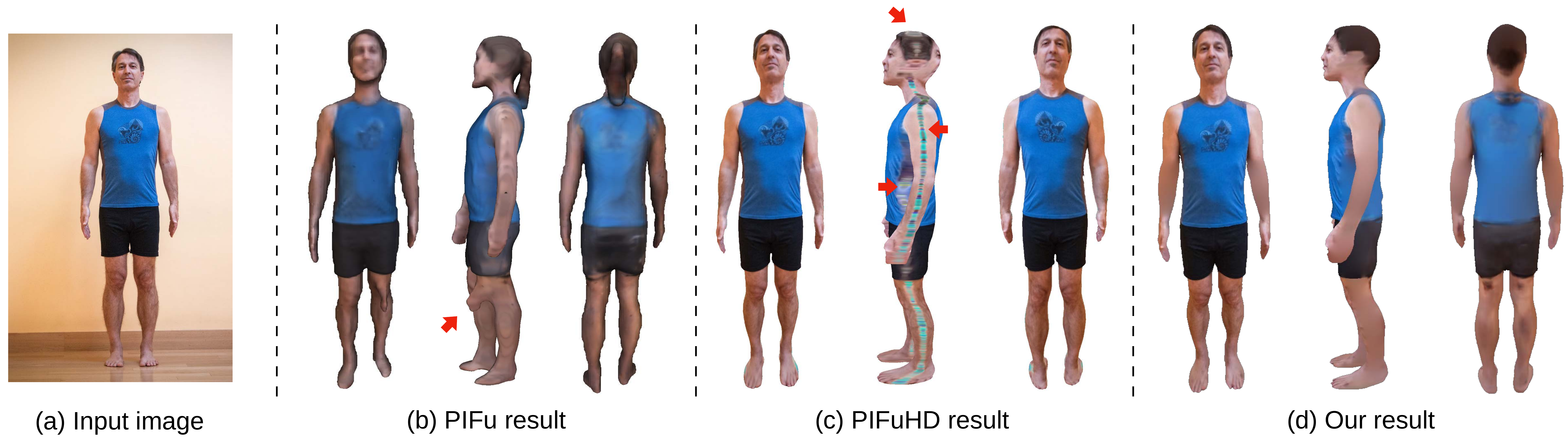}
	\caption{A comparison of reconstructing 3D characters from single images by the original methods of PIFu, PIFuHD, and our improved variant. Each 3D reconstruction result is shown in front, side, and back views. Salient errors are pointed by the red arrows. See text for details.}
	\label{fig:3d_model_comparison}
	\vspace{-0.3cm}
\end{figure*}

From a single 2D image, a 3D human character is extracted to preserve sufficient geometric and textural details consistent with the input. 
PIFu~\citep{saito2019pifu} and PIFuHD~\citep{saito2020pifuhd} are the two state-of-the-art methods on single-image based human shape recovery that have their unique pros and cons.
The 3D shapes and textures extracted by both methods are reasonably adhere to their 2D image inputs. Meanwhile, the texture map extracted by PIFu~\citep{saito2019pifu} has relatively low resolution and accuracy, see e.g. the protruded knee pointed by the red arrow in Fig.~\ref{fig:3d_model_comparison}(b). Although PIFuHD produces high-resolution 3D human geometry construction, notable errors are introduced at the unseen side by the symmetric assumption. As e.g. shown by the red arrows in Fig.~\ref{fig:3d_model_comparison}(c), the frontal human face is also erroneously synthesized at the back side of the 3D character head. 

Aiming at refining the reconstruction results, our improved variant takes advantage of PIFuHD in better estimating 3D geometry and camera-view appearance, as well as PIFu in better inpainting of texture for the unseen views. 
Moreover, we also adopt a heuristic in producing smooth transition near the boundary of visible and occluded surface regions, as follows:  
to detect the stitching boundary, we project the character (facing $Z_{+}$ direction) onto XY plane and match the edge of 2D silhouette with the 3D character; 
for a point $x$ in the transition region or inside the occluded region $O$ with color $c_x$, its color $c_x$ is expected to be close to the color $c_x^{\mathrm{p}}$ of the corresponding point on PIFu surface; at the same time, $c_x$ should also be close to those of its neighbors, $\mathcal{N}_x$. This is formulated as the following convex objective function, 
\begin{equation}
    \begin{aligned}
    \min \sum_{x \in O}\left[\|c_x - c_x^{\mathrm{p}}\|_2 + \lambda_{nn}\frac{1}{|\mathcal{N}_x|}\sum_{x'\in \mathcal{N}_x}\|c_x-c_{x'}\|_2 \right].
    \end{aligned}
    \label{eq:avatar_recovery}
\end{equation}
In practice, the vertex colors $c_x$ in $O$ are iteratively updated until a consistent convergence. For transition near the boundaries, only the second term of Eq.~\eqref{eq:avatar_recovery} is considered. As shown in Fig.~\ref{fig:3d_model_comparison}(d), our result is able to leverage the benefits of of both PIFu and PIFuHD methods, and produces a more natural transition near the boundary regions.

\subsubsection{Rigging, Animation, and Rendering}

\textbf{Fitting SMPL for extracted 3D shape.} 
The SMPL human shape, a generative 3D human representation controlled by pose and shape parameters, is used to facilitate the follow-up rigging and animation process. 
This requires to fit SMPL as close as possible to the reconstructed 3D human shape that amounts to estimating the pose ($\bm{\theta}$) and shape ($\bm{\beta}$) parameters by minimizing the following composite objective,
\begin{equation} 
\mathcal{L}(\bm{\beta}, \bm{\theta}) = \mathcal{L}_{\mathrm{surface}}(\bm{\beta}, \bm{\theta})  + \lambda_{j} \mathcal{L}_{\mathrm{joints}}(\bm{\beta}, \bm{\theta})  + \lambda_{r} \mathcal{L}_{\mathrm{reg}}(\bm{\theta}).
\label{eq:fit_smpl}
\end{equation}
The joints fitting term $\mathcal{L}_{joints}$ enforces the joints location of the SMPL shape to match with the predicted 3D joints from 2D image. Here, the initial 3D joints prediction $\hat{J}_{c}$ is obtained by regressing 2D joints from input image with OpenPose~\citep{cao2019openpose}, and by inverse projection into the reconstructed 3D human shape. 
Denote $f(\cdot)$ a transformation function of specific joint from initial position to current position following skeleton kinematics chain. 
Denote $\rho(\cdot)$ a differentiable Geman-McClure penalty function~\citep{geman1987statistical}, and $w$ the confidence of 2D joint prediction. We have,
\begin{equation}
\mathcal{L}_{\mathrm{joints}}(\bm{\beta}, \bm{\theta}) = \sum_{i\in|J|} \omega_i \rho \left( f\left( J(\bm{\beta})_i, \bm{\theta} \right)- \hat{J}_{c,i} \right).
\end{equation}
Then the surface fitting term $\mathcal{L}_{\mathrm{surface}}$ is applied to minimize distance between vertex $S^i$ of the reconstructed human shape $S$ and its nearest vertex $v$ of the SMPL shape $\mathcal{M}(\bm{\beta},\bm{\theta})$,
\begin{equation}
\mathcal{L}_{\mathrm{surface}}(\bm{\beta}, \bm{\theta})  =  \sum_{i \in |S|} \min_{v \in \mathcal{M}(\bm{\beta},\bm{\theta})} \left\|S^i -v\right\|_2.
\end{equation}
Finally, the pose regularization term $\mathcal{L}_{\mathrm{reg}}(\bm{\theta})$ penalizes unusual poses through the learned Gaussian mixture model from CMU dataset~\citep{cmu2003mocap}. Following~\citep{bogo2016keep}, it is of the form

\begin{equation}
\mathcal{L}_{\mathrm{reg}}(\bm{\theta}) = -\log\sum_{i}(g_i N(\bm{\theta}; \mu_{\bm{\theta},i},\Sigma_{\bm{\theta}, i})),
\end{equation}
where $N(\bm{\theta}; \mu_{\bm{\theta},i},\Sigma_{\bm{\theta}, i})$ is a Gaussian distribution with its mean $\mu_{\bm{\theta},i}$ and variance $\Sigma_{\bm{\theta}, i}$, and $g_i$ are weights of mixture Gaussian model.

In practice, to minimize the above objective function, during the first two iterations we only consider the joints and the pose regularization constraints for quick convergence; the surface constraint is then incorporated during the rest iterations. 

%


\textbf{3D model deformation and animation.} After obtaining the above optimized SMPL model that closely fits to the reconstructed 3D human mesh model, the SMPL model is used as an anchor to deform the 3D models to new poses. To start with, the vertex-level correspondences between the SMPL surface and the 3D human model are established by nearest neighbor search. In addition, body part information is used to eliminate possible mismatched pairs, especially these around the inter-joint of arms and torso. Specifically, the body parts information of reference image could be obtained using DensePose~\citep{alp2018densepose}, which then are back-projected to the surface of the 3D shape. As SMPL shape has pre-defined body segmentation, this could be utilized to filter out vertex pairs coming from different body parts. \revisions{Next, we compute a displacement map from the optimized SMPL mesh to their correspondences on the 3D human model, }
\begin{equation}
    S^j = \mathcal{M}_i(\bm{\beta}^{*},\bm{\theta}^{*}) + d_{i\rightarrow j}.
    \label{eq:disp}
\end{equation}
\revisions{where $\bm{\beta}^{*}$ and $\bm{\theta}^{*}$ are the optimized shape and pose parameters of the SMPL model. $S_j$ and $\mathcal{M}_i(\bm{\beta}^{*},\bm{\theta}^{*})$ are the correspondences and $d_{i\rightarrow j}$ is the displacement from optimized SMPL model to reconstructed 3D human model. }

\revisions{Intuitively, to repose the human shape, we could acquire the target positions $S^{*}$ of shape vertices by applying the displacement map to the reposed SMPL as in Eq.\eqref{eq:disp}. However, this will lead to imperfections due to free-form deformation. Following~\cite{zuo2020SparseFusion}, we instead utilize the vertices of $S^{*}$ as control points to deform the 3D human model as rigid as possible, by enforcing a local rigidity constraint. The locally rigid deformation $\bm{R}$ and the deformed human model $\hat{S}$ are obtained by minimizing the following objective, }
\begin{equation}
    \begin{aligned}
    \mathcal{L}_{def}(\bm{R},\hat{S}) &= \sum_{i\in|S|}\sum_{j \in \mathcal{N}_i} k_{ij} \left\|(\hat{S}^i -\hat{S}^j) - \bm{R}_i(S^i - S^j)\right\|_2\\
    & + \sum_{l\in|S|}\left\|\hat{S}^l -S^{*,l}\right\|_2.
    \end{aligned}
\end{equation}
\revisions{Here $\mathcal{N}_i$ is the set of the neighboring vertices of $S^i$; $k_{ij}$ is the corresponding weights of neighboring vertices. $\bm{R}_i$ is a rotation matrix. The above objective function is optimized by iteratively solving the rotation matrix $R$ and the deformed mesh $\hat{S}$~\citep{sorkine2007rigid}.}

\textbf{Rendering.} The target 3D shape are deformed and driven by the generated pose sequences frame-by-frame, which are subsequently fed into 3D game engine (Unity3D) to integrate physical conditions such as illuminations and shadows and produce the final videos. Specifically, spot light and directional light are used to illuminate the character from top. Four cameras, fixed at half height of the 3D character, are aimed at the subject to record the \textit{front, back, left side} and \textit{right side} views, respectively. 

\section{Empirical Evaluations}
\label{experiments}

A comprehensive set of experiments are conducted to systematically evaluate the performance of our action2video approach, which consists of the two-step pipeline of action2motion and motion2video. 
We start by introducing the related datasets, and our implementation details. 
This is followed by a detailed examination of our action2motion process at Sec.~\ref{motionGeneration}, and comparisons for our motion2video with related efforts at Sec.\ref{subsec:motion2video}.
Finally, Sec.~\ref{subsec:action2video} provides a holistic evaluation of our full pipeline, action2video. 

\noindent{\bf Datasets. }
Ideally, we expect to work with motion datasets that contain considerable amount of distinct motion clips of various action categories, and with proper 3D pose annotations. 
In practice, we achieve this by postprocessing existing popular datasets, including re-annotating 3D positions of NTU-RGBD~\citep{shahroudy2016ntu} and action categories of CMU MoCap~\citep{cmu2003mocap}. We also curate an in-house dataset, \DN. In these three datasets, all human poses are uniformly annotated into 3D joints connected into 5 kinematics chains, with pelvis being the root joint.

\begin{itemize}
    \item \textbf{NTU-RGBD} is a large-scale 3D human motion dataset containing nearly one million motion sequences of 120 action types. Its pose annotation (i.e. 3D joint positions) is from MS Kinect readout, which is known unreliable and temporally unstable. In our experiments, the state-of-art video 3D shape estimation method~\citep{kocabas2019vibe} is employed to re-estimate the 3D poses from video feeds. 
    Note in our scenario, it's sufficient for these poses to appear realistic, and they are not necessarily matched perfectly with the true poses. 
    A subset of 13 distinct actions are further selected in our empirical evaluation, such as \textit{cheer up, pick up, salute}, consisting of 3,900 motion clips. Each pose is represented by 18 joints (i.e. 17 bones).

    \item \textbf{CMU MoCap} is dataset accurately annotated by motion capture markers, with 2,605 pose sequences. However, the dataset is not originally organized by action types. We identify 8 distinct actions based on their motion captions, including \textit{running, walking, climbing, jumping}. In the end, 1,088 motions are re-organized by action type, with each skeleton constituting 22 3D joints (i.e. 21 bones). In implementation, these pose sequences are down-sampled from 100 HZ to a frequency of 12 HZ.

    \item \textbf{\DN} is our in-house dataset 
    that comes with proper annotations. It consists of 1,191 motion clips and 90,099 frames in total, which are categorized into 12 coarse-grained action categories, including e.g. \textit{warm up, lift dumbbell}, and 34 fine-grained action types such as \textit{warm up (Leg pressing), lift dumbbell (with right hand)}. The fine-grained annotations give more specific and dedicated information of the motions. We test our model on both coarse- and fine-grained annotations. Our dataset, \DN, contains more accurate and stable 3D position annotations compared to NTU-RGBD; and has more well-organized action annotations than CMU MoCap. Note each body pose contains 24 joints (i.e. 23 bones).
\end{itemize}

\begin{figure}[tb]
	\centering
	\includegraphics[width=\linewidth]{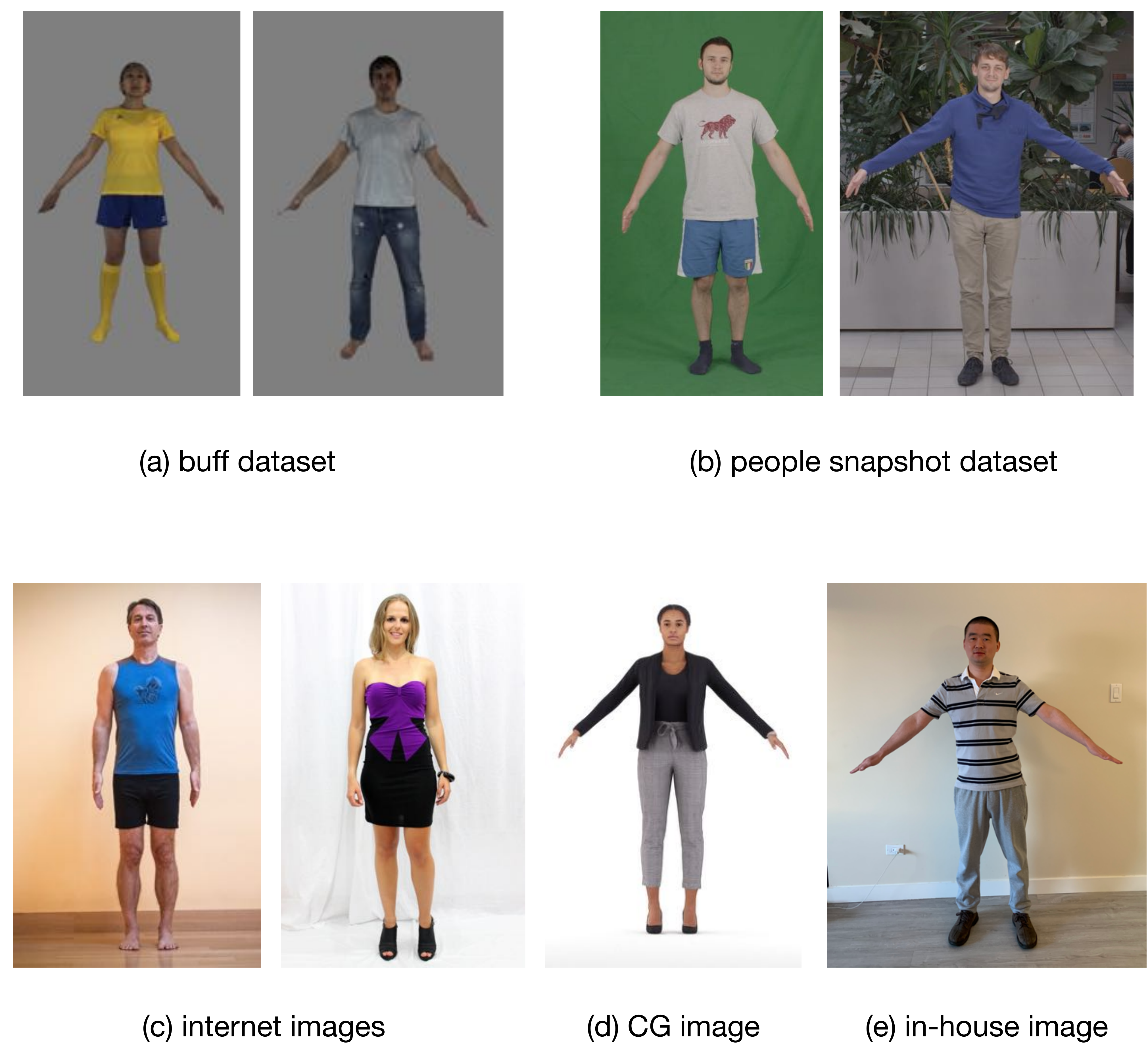}
	\caption{\revision{Input images used in our experiments are from different sources, including (a) BUFF dataset~\citep{Zhang_2017_CVPR}, (b) People Snapshot dataset~\citep{alldieck2018video}, (c) internet images, (d) CG image, and (e) our in-house captured images. See text for details.}}
	\label{fig:image_source}
	\vspace{-0.3cm}
\end{figure}

\revision{To showcase that our pipeline could work with wide range of applications, input images from myriad sources are considered in our experiments, as displayed in Fig.~\ref{fig:image_source}. They include images from the BUFF dataset~\citep{Zhang_2017_CVPR}, People Snapshot dataset~\citep{alldieck2018video}, as well internet images, computer-generated (CG) images~\footnote{https://renderpeople.com/3d-people/}, and our in-house captured images. BUFF dataset provides 26 4D human sequences with different cloth styles and performing different actions. We then render 2D images from these human shapes. People Snapshot dataset contains 12 subjects and 24 video sequences with different backgrounds. More examples are provided in the supplementary file.}

\noindent{\bf Implementation Details. }
Our action2video pipeline is mostly implemented by PyTorch. For all encoder layers, the output size is set to 128. One-layer GRU is used for prior network, posterior network and pose decoding module, while generator uses two-layer GRU. The hidden unit size of GRU is 128. And the noise vector $\mathbf{z}$ and $\mathbf{h}_t^o$ has the dimension of 30 and 20 respectively. The Adam optimizer is applied for training throughout all experiments, with learning rate of 0.0002, weight decaying of 0.00001, and default parameter values including $\beta_1=0.9$, $\beta_2=0.999$. Our model is trained with mini-batch size of 128. To stabilize the training process, \textit{teacher forcing rate} $p_{\mathrm{tf}}$ is set to 0.6. The values of aforementioned hyper-parameters are fixed throughout our empirical experiments across all datasets.

Afterwards, we generate motions with length of 60, 100 and 60 on NTU-RGBD, CMU MoCap and \DN, respectively. The hyper-parameter $\lambda_{kl}$ is a trade-off between reconstruction constraints and KL-divergence penalty. During training, the value of $\lambda_{kl}$ for all datasets are initialized with 0.001 and linearly increased to 0.1, 0.1 and 0.01 at the end for above datasets respectively. During training, the value of $\lambda_{align}$ is set to 10 throughout these experiments.

In motion2video step, to extract 3D shape from single image, $\lambda_{nn}$ and 10 neighbors are used in Eq.~\eqref{eq:avatar_recovery} for occluded region. The values of $\lambda_j$ and $\lambda_r$ in Eq.~\eqref{eq:fit_smpl} are set to 2.0 and 0.2,  respectively.

\subsection{Step 1: Action2motion} 
\label{motionGeneration}

Thorough evaluations of the action2motion step are carried out in this section. They include both quantitative and qualitative reports of motion generation results, and fine-grained analysis of the locomotion generation module;
We also provide demonstrations of specific action2motion applications such as motion interpolation in the latent space, motion transition, and 3D motion \revision{outpainting}.
By default, the action2motion GLMI-M variant is utilized in our approach.

\subsubsection{Evaluations}\label{subsec:evaluations}

We start by introducing a tally of evaluation metrics and baseline methods used throughout this section, which is followed by a series of qualitative and quantitative evaluations.

\noindent{\bf Evaluation Metrics. } \label{subsubsec:metrics}We aim to evaluate the generated motions from the aspects of being \textit{natural} and \textit{diverse}. To achieve this, the three metrics in~\cite{lee2019dancing} are adopted in our evaluations: \textit{Frechet Inception Distance(FID)} to characterize the visually realistic aspect, \textit{Diversity} and \textit{Multimodality} to quantify the diverse levels. The \textit{action recognition accuracy} is additionally used to gauge the similarity between generated motions and real-life motions, as well as the degree of generated motions belonging to the prescribed action. 

FID is perhaps the most important indicator in our scenario. A \textit{lower} FID suggests a better result. For multimodality and diversity, a result is claimed better only if its diversity and multimodality scores are \textbf{closer} to their respective values obtained from real motions. To calculate these metrics, we rely on a feature extractor to obtain the high-level features of motions. \revision{Since there is no standard implementation of such motion feature extraction, a vanilla RNN action recognition classifier is trained for each dataset; and the final layer of classifier is used as the motion feature extractor.} 

We elaborate these four metrics as below:
\begin{itemize}[leftmargin=*]
    \item \textbf{Frechet Inception Distance}(FID): FID is an effective metric to evaluate the overall quality in motion generation. A large amount (in our case, 3,000) of generated motions and real motions are sampled and then are transformed to two sets of features. For real motion, we sample from test set with replacement. Then, FID is measured by computing the distance between the feature distribution of generated motions and that of the real motions. 
    
    \item \textbf{Recognition Accuracy}: Recognition accuracy is calculated as the accuracy of applying a pre-trained RNN action recognition classifier to the motion of interest.
    
    \item \textbf{Diversity}: Diversity indicates the variance of the motions across \textit{all} action types. Specifically, a large set of motions are sampled from all varieties of action types, from which two subsets are randomly sampled with the same size $S_d$. The corresponding sets of motion feature vectors $\{\mathbf{v}_1,\ldots,\mathbf{v}_{S_d}\}$ and $\{\mathbf{v}_1',$ $\ldots,\mathbf{v}_{S_{d}}'\}$ are extracted respectively. Then, the diversity of this set of motions is evaluated by
    \begin{equation}
        \mathrm{Diversity} = \frac{1}{S_d}\sum_{i=1}^{S_d}\parallel \mathbf{v}_i-\mathbf{v}_i' \parallel_2,
    \end{equation}
    where $S_d=200$ is used throughout our experiments.
    
    \item \textbf{Multimoldality}: Different from diversity, multimodality indicates how much the sampled motions vary within \textit{each} action category. Suppose there are $C$ action types in the set of motion sequences. For the $c$-th action, two subsets with same size $S_m$ are randomly sampled, which are then transformed to two subset of feature vectors $\{\mathbf{v}_{c,1},\ldots$ $,\mathbf{v}_{c,S_m}\}$ and $\{\mathbf{v}_{c,1}',\ldots,\mathbf{v}_{c,S_m}'\}$. The multimodality is defined as
    \begin{equation}
        \mathrm{Multimodality} = \frac{1}{C \times S_m} \sum_{c=1}^C \sum_{i=1}^{S_m} \left\| \mathbf{v}_{c,i}-\mathbf{v}'_{c,i} \right\|_2,
    \end{equation}
    where ${S_m}=20$ is used in our experiments.
\end{itemize}

\begin{figure*}[t]
  \centering
    \begin{tabular}{c c}
      \qquad\qquad \underline{\tiny{\textbf{Click}$\downarrow$}} & \underline{\tiny{\textbf{Click}$\downarrow$}} \\
      \qquad \animategraphics[height=0.8cm]{6}{figures/qualitative_visual/q_0_}{0}{59}  &
      \animategraphics[height=0.8cm]{6}{figures/qualitative_visual/q_1_}{0}{59} \\
      \multicolumn{2}{c}{
      \includegraphics[width=\linewidth]{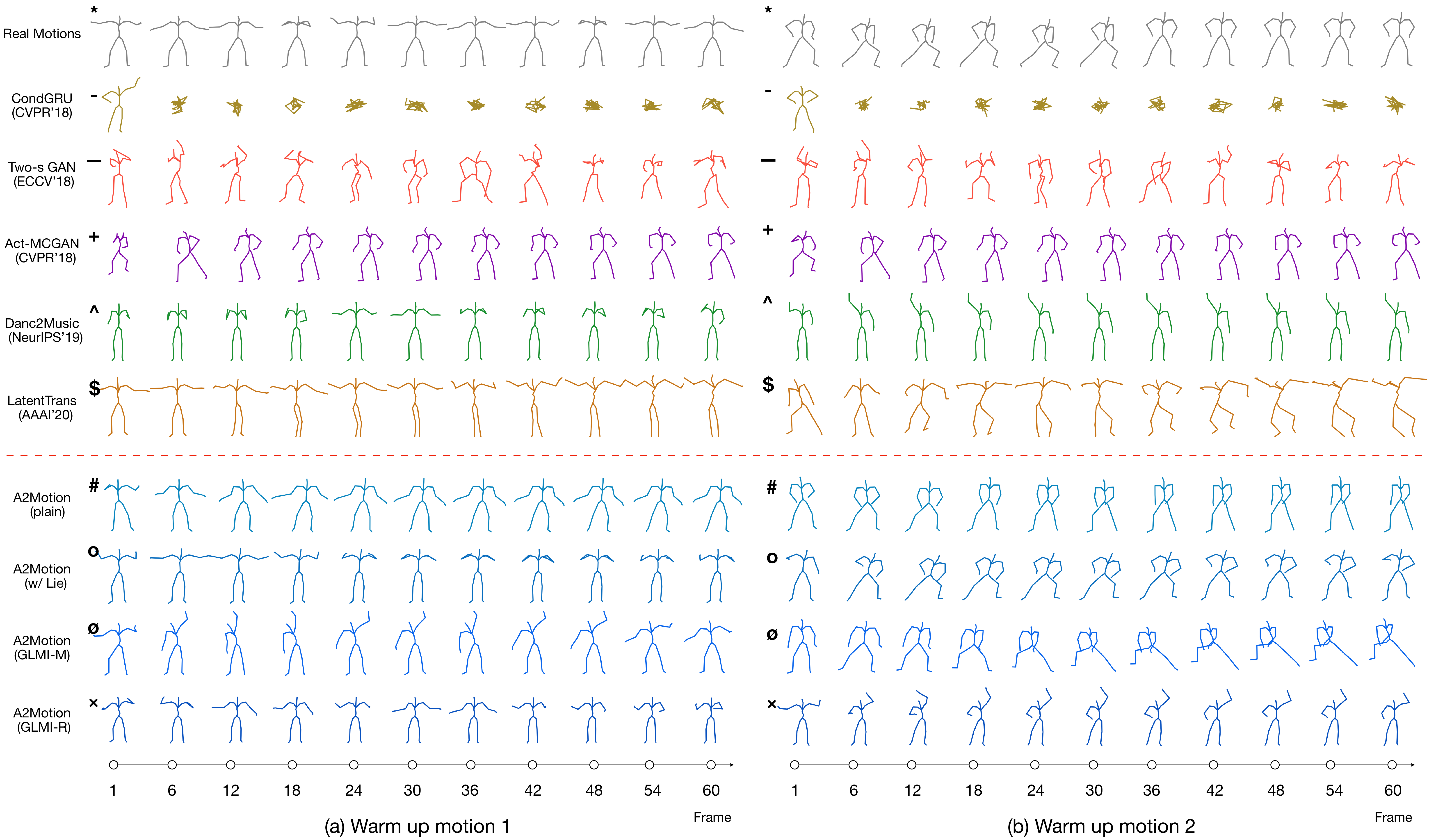}
      }
  \end{tabular}
  \caption{Visual comparison of motions generated by the baseline methods and our four action2motion variants. Two \textit{warm up} motion sequences are sampled for each of the comparison methods. Every 6th frame is shown. See text for details. Best viewed in Adobe Acrobat Reader to activate the animations by clicking the boxed items in the top row. Note each item in the top row is with specific tag and color corresponding to its row of motion sequence displayed below.
  }
  \label{fig:qualitative_result1}
  \vspace{-0.3cm}
\end{figure*}

\noindent{\bf Baseline methods. } 
Since the problem of action2motion, aka action-conditioned 3D human motion generation, is relatively new, there are few existing methods to compare with. 
We thus adapt the state-of-art methods from related areas to our context, as follows:
\begin{itemize}[leftmargin=*]
    \item \textbf{CondGRU}. Condition GRU is used as a deterministic baseline in our setting, which is also the principal model for audio-to-motion translation in~\cite{shlizerman2018audio} and text-to-motion generation in~\citep{ahn2018text2action,stoll2020text2sign}. Here, a small modification of the model is made that the input is the concatenation of condition vector and pose vector at present step and the output is the pose vector for next step.
    
    \item \textbf{Two-stage GAN}. \cite{cai2018deep} propose a two-stage GAN method for 2D human motion generation based on action types. In particular, a Wasserstein GAN~\citep{arjovsky2017wasserstein} is first trained as the pose generator. After that, the motion generator is learned to produce input latent vector for pose generator to synthesize pose at each time. By using adversarial training, the entire generated pose sequences are judged by a motion discriminator. We adapt this method for 3D human motion generation through necessary modifications.
    
    \item \textbf{Act-MoCoGAN}. MoCoGAN~\citep{tulyakov2018mocogan} is a widely used method for both conditional and unconditional video generation. While generating a video, the input noise vector are composed of two parts: one is a shared vector over time, another is a instinct noise vector sampled at each time. These two inputs are expected to map to the stationary content and dynamic motions in videos. In our experiment, to generate 3D human dynamics, we keep the original architecture and replace the video and image discriminators to motion and pose discriminators, respectively.
    
    \item \revision{\textbf{Dancing2Music}. Dancing2Music~\citep{lee2019dancing} generates 2D dancing motion sequences from audio signals, which consists of two main stages, decomposition and composition. During decomposition, a motion sequence is segmented into short motion snippets, with dance unit VAE (DU-VAE) model being trained to generate these motion snippets given the latent vectors of motion content and an initial frame; during composition, a music-to-movement GAN (MM-GAN) is trained to generate latent vectors of motion snippet contents conditioned on the given music signals. To make a meaningful comparison, the official implementation is adapted by replacing the music signals with action categories}.
    
    \item \revision{\textbf{LatentTransition}. \cite{wang2020learning} consider a two-stage GAN~\citep{cai2018deep}, with a Bi-LSTM being employed to produce input latent vectors for pose generation. An additional auxiliary action classifier further ensures the action-awareness of the generative model.}
    
    \item \textbf{Action2Motion (plain)}. Oue action2motion variant by adopting the pose decoding module of Fig.~\ref{fig:pose_decoding}(a), where the 3D position of joints are directly produced from generator.
    
    \item \textbf{Action2Motion (w/ Lie)}. Our action2motion variant with the pose decoding module of Fig.~\ref{fig:pose_decoding}(b), where the Lie algebra parameters and root joint locations are generated independently. 
    
    \item \textbf{Action2Motion (GLMI-M)}. Our action2motion variant with the pose decoding module of Fig.~\ref{fig:pose_decoding}(c), where both the Lie algebra and GLMI are used, and GLMI is implemented by MLP.
    
    \item \textbf{Action2Motion (GLMI-R)}. Our action2motion variant with the pose decoding module of Fig.~\ref{fig:pose_decoding}(d), where both the Lie algebra and GLMI are used, and GLMI is implemented by GRU network instead.
\end{itemize}

\begin{figure*}[tb]
  \centering
    \begin{tabular}{c c}
        \underline{\tiny{\textbf{Click}$\downarrow$}} & \underline{\tiny{\textbf{Click}$\downarrow$}} \\
      \qquad \qquad
      \animategraphics[height=0.95cm]{6}{figures/fine_grained/0_}{0}{59} 
      \qquad \qquad&
      \animategraphics[height=0.95cm]{6}{figures/fine_grained/1_}{0}{59} \\
       \multicolumn{2}{c}{
    \includegraphics[width=0.99\linewidth]{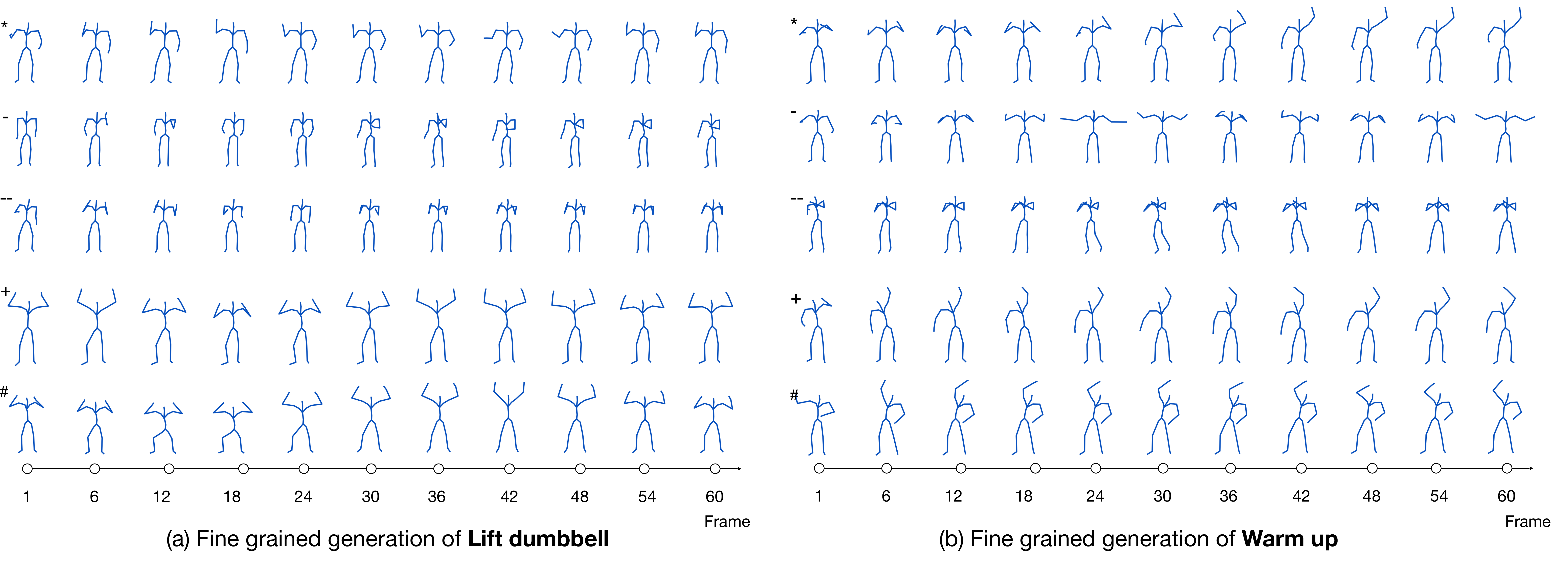}
    }
     \end{tabular}
    \caption{Motion examples of fine-grained action categories generated by our action2motion (GLMI-M). Every 6th frame is shown. (a) Lift dumbbell with (from top to bottom) \textit{right hand, left hand, both hand, both hand over head}, and \textit{both hand over head and squat}. (b) Warm up with (from top to bottom) \textit{alt chest expansion, chest expansion, wrist circles, left side reach} and \textit{right side reach}.
    Best viewed in Adobe Acrobat Reader to activate the animations by clicking the boxed items in the top row. Note each item in the top row is with specific tag corresponding to its row of motion sequence displayed below.
    }
    \label{fig:fine-grained}
\end{figure*}

\noindent{\bf Visual comparisons. }\label{visualComparisons}
Fig.~\ref{fig:qualitative_result1} provides qualitative comparisons of skeletal motions generated from different methods: given an action category of \textit{warm up}, two motions of length 60 are sampled, with every 6th frame being displayed. 

Conditional GRU~\citep{shlizerman2018audio} requires as input an initial ground-truth pose to kick-start its generation process. Unfortunately the generated poses often collapse into a cloud of 3D points near the root joint. 
Two-stage GAN~\citep{cai2018deep} produces better results, which however are still perceptually not satisfactory. 
The skeletal sequence result of Act-MoCoGAN by~\cite{tulyakov2018mocogan} is visually the best among these three methods. 
The generated poses nonetheless often froze to a fixed posture quickly. \revision{Dancing2Music~\citep{lee2019dancing} shows capability of yielding natural poses and motions. Meanwhile, a single such motion usually contains multiple actions, with the motion context deviating from the prescribed action type. For instance, in the left column of Fig.~\ref{fig:qualitative_result1}, the stick man first performs \textit{lift dumbbell} (from $t = 1$ to $t = 18$), then a short-time \textit{warm up} (from $t = 24$ to $t = 36$), and finally drifting into \textit{drinking}. On the other hand, LatentTransition~\citep{wang2020learning} always starts with natural poses, then struggles with proper modeling of long-term motion dependencies, which typically deteriorates to unrecognizable movements. }
These results are in sharp contrast to that of our four action2motion variants, whose results are in general visually more appealing.
Here, the action2motion (plain) variant sometimes generate visual defects noticeable to human eyes. 
For example, in the left column of Fig.~\ref{fig:qualitative_result1}, the arm bone lengths of the same individual abnormally vary from $t=1$ to $t=24$. 
This is due to the intrinsic 3D-coordinate skeletal representation adopted by the plain variant that does not obey the underlying skeletal kinematics. 
Skeletal motions generated by the other action2motion variants are typically more faithfully resemble to real-life motions, which we attribute to their adherence to kinematics by their use of Lie group/algebraic skeletal representations. 

\revision{Diversity is another important evaluation criteria. In Fig.~\ref{fig:qualitative_result1}, motions generated from conditional GRU tends to be visually least appealing; this is followed by those of two-stage GAN and LatentTransition;
the results of Act-MoCoGAN often suffers from the \textit{mode collapsing} issue, with similar results popping up after multiple separate runs; In comparison, Dancing2Music is capable of producing diverse motions by transiting between different short motion snippets. However, the generated motions could not be faithfully aligned to the prescribed action type; On the contrary, our action2motion variants are shown to be capable of generating both diverse and consistent motions.} 

Moreover, our action2motion framework is also capable of producing motions from fine-grained action categories, as showcased in Fig.~\ref{fig:fine-grained}. 
The motions generated by our action2motion (GLMI-M) variant faithfully assemble the subtle characteristics of local motions (e.g. leg pressing and chest expansion), and body parts (e.g. left hand and right hand) from a range of fine-grained action types. 


\begin{table*}[t]
  \caption{Performance evaluation on HumanAct12 benchmark on coarse-grained and fine-grained action categories, respectively. $\pm$ indicates 95\% confidence interval. $\uparrow$ (or $\downarrow$) is higher (or lower) the better;  $\rightarrow$ means closer to real motion scores the better. For performance, \textbf{bold} face specifies the best method, with underscore referring to the second best.}
  \vspace{-5pt}
  \label{tab:performance1}
  \resizebox{\textwidth}{60pt}{
    \begin{tabular}{l c c c c c c c c c c c}
    \toprule
    \multirow{2}{*}{Methods} & \multicolumn{4}{c}{\DN (Coarse-grained)} & & \multicolumn{4}{c}{\DN (Fine-grained)} \\
    \cline{2-5}
    \cline{7-10}
                    & FID$\downarrow$ & Accuracy$\uparrow$ & Diversity$\rightarrow$& MModality$\rightarrow$ &  & FID$\downarrow$ & Accuracy$\uparrow$ & Diversity$\rightarrow$ & MModality$\rightarrow$\\   
    \midrule

            \textbf{Real motions}    & \et{0.092}{.007}  &  \et{0.997}{.001}& \et{6.853}{.053}& \et{2.449}{.038}  & & \et{0.133}{.004}  &  \et{0.991}{.001}& \et{7.001}{.018}& \et{2.666}{.012}\\
    \midrule
            CondGRU   & \et{40.61}{.144}  &  \et{0.080}{.002}& \et{2.381}{.020}& \etb{2.341}{.036}  & & \et{33.91}{.059}  &  \et{0.034}{.001}& \et{3.779}{.034}& \et{3.469}{.026} \\
            Two-stage GAN    & \et{10.48}{.089} &  \et{0.421}{.006}& \et{5.960}{.049}& \ets{2.805}{.036}  & & \et{6.956}{.038}  &  \et{0.397}{.002}& \et{6.151}{.017}& \etb{2.694}{.008} \\
            Act-MoCoGAN   & \et{5.610}{.113}  &  \et{0.793}{.004}& \etb{6.752}{.071}& \et{1.055}{.017}  & & \et{2.468}{.026}  &  \etb{0.832}{.002}& \ets{6.891}{.023}& \et{0.878}{.003}\\
            
            Dancing2Music & \et{3.832}{.103} & \et{0.145}{.003} & \et{6.523}{.096}&
            \et{6.313}{.035} & &\et{3.484}{.085} & \et{0.029}{.001} & \et{6.567}{.106} &\et{6.406}{.026} \\
            LatentTransition & \et{3.553}{.093} &\et{0.471}{.005} & \et{6.580}{.110} &
            \et{4.387}{.039} & &\et{2.123}{.044} &\et{0.397}{.004} &\et{6.640}{.082} &\et{4.590}{.027} \\
            
    \midrule
            \textit{\textit{Action2Motion}} (plain)    & \et{3.299}{.079}  &  \et{0.656}{.005}& \ets{6.742}{.046}& \et{4.248}{.037}  & & \et{1.329}{.021}  &  \et{0.560}{.002}& \et{6.756}{.015}& \et{4.487}{.015}\\
            \textit{Action2Motion} (w/ Lie)    & \et{2.458}{.079}  &  \etb{0.923}{.002}& \et{7.032}{.038}& \et{2.870}{.037}  & & \et{1.000}{.016} &  \et{0.776}{.001}& \et{6.783}{.015}& \et{3.508}{.011}\\
            \textit{Action2Motion} (GLMI-M)    & \etb{2.157}{.052}  &  \ets{0.835}{.005}& \et{6.986}{.028}& \et{3.633}{.031}  & & \etb{0.739}{.015} &  \ets{0.787}{.002}& \et{6.783}{.015}& \ets{3.301}{.009}\\
            \textit{\textit{Action2Motion}} (GLMI-R)    & \ets{2.349}{.057}  &  \et{0.831}{.002}& \et{7.001}{.023}& \et{3.607}{.037}  & & \ets{0.957}{.017}  &  \et{0.767}{.001}& \etb{6.924}{.019}& \et{3.303}{.012}\\
    \bottomrule
  \end{tabular}
  }
  \vspace{-0.3cm}
\end{table*}

\begin{table*}[htb]
  \caption{Performance evaluation on CMU MoCap and NTU-RGBD Dataset. $\pm$ indicates 95\% confidence interval. As NTU-RGBD dataset does not have global motion trajectory annotations available, our GLMI-M \& GLMI-R variants that could not be fairly evaluated here. $\uparrow$ (or $\downarrow$) is higher (or lower) the better;  $\rightarrow$ means closer to real motion scores the better.  For performance, \textbf{bold} face specifies the best method, with underscore referring to the second best.}
  \label{tab:performance2}
  \resizebox{\textwidth}{60pt}{
    \begin{tabular}{l c c c c c c c c c c c}
    \toprule
    \multirow{2}{*}{Methods} & \multicolumn{4}{c}{CMU MoCap} & & \multicolumn{4}{c}{NTU-RGBD} \\
    \cline{2-5}
    \cline{7-10}
                    & FID$\downarrow$ & Accuracy$\uparrow$ & Diversity$\rightarrow$& MModality$\rightarrow$ &  & FID$\downarrow$ & Accuracy$\uparrow$ & Diversity$\rightarrow$ & MModality$\rightarrow$\\   
    \midrule

            \textbf{Real motions}    & \et{0.064}{.006}  &  \et{0.936}{.002}& \et{6.130}{.079}& \et{2.726}{.066}  & & \et{0.031}{.004}  &  \et{0.999}{.001}& \et{7.108}{.048}& \et{2.194}{.025}\\
    \midrule
            CondGRU   & \et{51.72}{.123}  &  \et{0.093}{.001}& \et{0.792}{.011}& \et{0.752}{.016}  & & \et{28.31}{.138}  &  \et{0.078}{.001}& \et{3.663}{.024}& \et{3.578}{.027} \\
            Two-stage GAN    & \et{14.34}{.107} &  \et{0.179}{.003}& \et{4.419}{.064}& \etb{1.623}{.024}  & & \et{13.86}{.091}  &  \et{0.202}{.003}& \et{5.328}{.039}& \et{3.490}{.027} \\
            Act-MoCoGAN   & \et{11.15}{.074}  &  \et{0.445}{.005}& \et{5.280}{.069}& \ets{1.516}{.022}  & & \et{2.723}{.019}  &  \etb{0.997}{.001}& \et{6.920}{.061}& \et{0.907}{.009}\\
            Dancing2Music & \et{6.882}{.127} & \et{0.138}{.003} & \et{4.772}{.104} &\et{4.289}{.012} & & \et{3.461}{.077} & \et{0.075}{.002} & \et{6.562}{.114} &\et{6.556}{.045}\\
            LatentTransition & \et{12.85}{.181} &\et{0.389}{.003} &\et{5.856}{.143} &\et{4.639}{.053} & &\et{6.882}{.127} &\et{0.138}{.003} &\et{4.772}{.105} &\et{4.289}{.049}\\
    \midrule
            \textit{Action2Motion} (plain)    & \et{2.994}{.052}  &  \et{0.378}{.004}& \ets{5.791}{.044}& \et{5.006}{.045}  & & \ets{0.540}{.047}  &  \et{0.832}{.004}& \ets{6.926}{.049}& \ets{3.443}{.052}\\
            \textit{Action2Motion} (w/ Lie)   & \et{2.885}{.116}  &  \etb{0.686}{.003}& \et{6.509}{.061}& \et{4.126}{.056}  & & \etb{0.330}{.008} &  \ets{0.949}{.001}& \etb{7.065}{.043}& \etb{2.052}{.030}\\
            \textit{Action2Motion} (GLMI-M)   & \etb{2.448}{.031}  &  \et{0.665}{.001}& \etb{6.374}{.022}& \et{4.093}{.019}  & & - &  - & - & -\\
            \textit{Action2Motion} (GLMI-R)     & \ets{2.519}{.029}  &  \ets{0.675}{.001}& \et{6.484}{.028}& \et{4.073}{.029}  & & -  & - & - & -\\
    \bottomrule
  \end{tabular}
  }
  \vspace{-0.2cm}
\end{table*}

\noindent{\bf Quantitative comparisons. }Quantitative evaluations are conducted on a range of datasets. Specifically, Table~\ref{tab:performance1} displays results on our in-house HumanAct12 dataset, where coarse-grained and fine-grained action annotations are both considered; Table~\ref{tab:performance2} presents comparison results on the popular benchmarks of CMU MoCap and NTU-RGBD. Considering the stochastic nature of motion generation, each experiment is repeated 20 times, a statistical confidence interval of 95\% is reported in both tables. 
Note action2motion (GLMI) is however not applicable to the post-processed NTU-RGBD dataset, since the re-estimated pose sequences from videos does not contain global trajectory information.

Among the four evaluation metrics in both tables, FID is perhaps the most important indicator, as it evaluates the overall quality of the generated motions. 
Recognition accuracy quantifies how well a generated motion fits into an action category. Diversity and multimodality (i.e. MModality) are metrics quantifying the diversity aspects of the generated motions.
Note the values of FID (or accuracy) is lower (or higher) the better; for Diversity and MModality though the values are as close to the real motion scores the better. 
From Table~\ref{tab:performance1} and Table~\ref{tab:performance2}, we have the following observations. 
As a deterministic method, conditional GRU fails to generate diverse motions that is essentially an one-to-many mapping problem. 
\revision{GAN models such as two-stage GAN, Act-MoCoGAN and LatentTransition have improved upon conditional GRU in both metrics of FID and recognition accuracy. The considerably high accuracy obtained by Act-MoCoGAN may be attributed to its use of action classifier during training. A sharp drop of FID is observed in Dancing2Music, which however comes at the price of much lower accuracy.} 
Meanwhile, our action2motion clearly outperforms the rest on FID, and the GLMI-M variant consistently excels among the four action2motion variants. 
The success could be partly attributed to the incorporation of Lie algebraic pose representation. 
%

Given substantial performance on FID and perhaps also accuracy scores, the scores of diversity and multimodality are also important indicators for the model capacity of producing diverse motions. 
Note for diversity and multimodality, the higher values do not necessarily reflect better performance; instead the values are best to be close to those from the real motions, denoted as $\rightarrow$ in Tables~\ref{tab:performance1} and~\ref{tab:performance2}. 
Act-MoCoGAN generates motions with severely limited diversity. Overall, our action2motion variants, while performing best on FID and accuracy, also maintain a considerable extent of diversity and multimodality. 

\noindent{\bf Crowd-sourced Subjective Evaluation. }\label{userStudy}
In addition to the aforementioned objective experiments, two user studies are conducted on Amazon Mechanical Turk. The principal criteria used in these two user surveys are the visual perceptual quality of the motion, and the magnitude it is adhere to the intended action categories. Users who possess hit approval rate higher than 97\% and 1000 completed hits are considered.

\begin{figure}[tb]
  \centering
  \includegraphics[width=\linewidth]{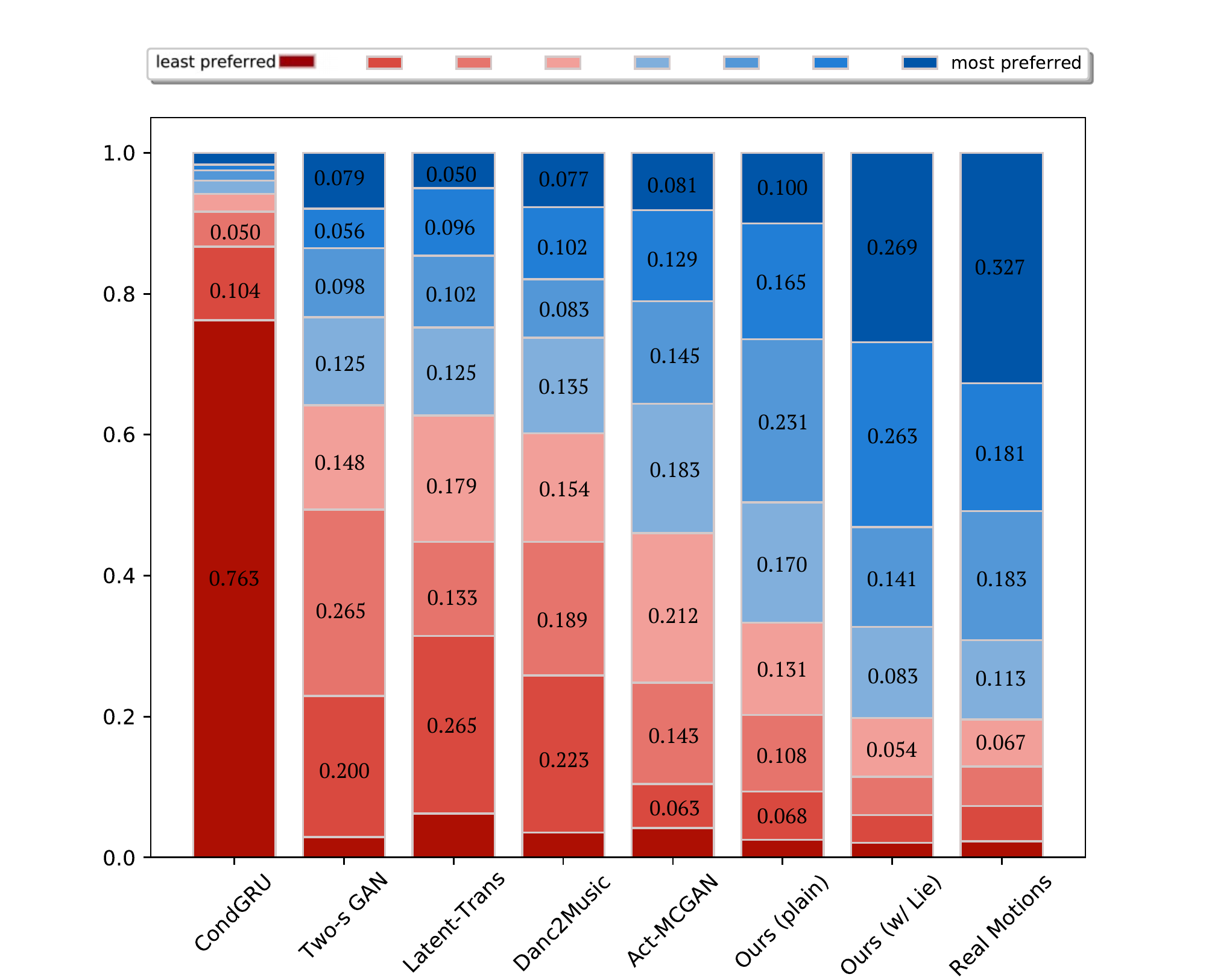}
  \setlength{\abovecaptionskip}{0.1cm}
  \setlength{\belowcaptionskip}{-0.2cm}  
  \vspace{-7pt}
  \caption{Crowd-sourced subjective assessment results of motions generated by comparison methods. For each method, there is a bar of different colors (from red to blue) indicating the percentage of corresponding preference levels (least to most preferred). See text for details.}
  \label{fig:rank_hist}
  \vspace{-0.3cm}
\end{figure}

The first user study is illustrated in Fig.~\ref{fig:rank_hist}, which compares the first two action2motion variants, ours (plain) and ours (w/ Lie), with baseline methods. 
Here, same amount (i.e. 36) of motions are generated by different methods. The users are then asked to rank their preferences of these motions evenly sampled over all action categories. 
Our action2motion variants receive the highest user ratings. \revision{Contrarily, conditional RNN, two-stage GAN and LatentTransition are the three least performed methods. Dancing2Music and Act-MoCoGAN rank somewhere in-between. More positive feedback is observed in our action2motion \textit{plain} variant, with 10\% motions being graded the first by users. By adopting the Lie algebraic representation, our ours w/ Lie variant further narrows the gap to real motions, with 54\% generated motions being secured at the top-2 spots by user ratings.}

The second user study compares bewteen our two action2motion variants: ours (GLMI) and ours (w/ Lie). 
As GLMI-M outperforms GLMI-R in most cases, we focus on the evaluation of GLMI-M in this survey. 
Here the motions are generated following the same protocol conceived in the first study. 
As shown in table~\ref{tab:rank_over}, ours with GLMI earns more appreciation from users when compared with ours (w/ Lie), with over a half motion sequences (i.e. 54.4 \%) being preferred by users. When comparing to real motions, samples generated by ours (w/ Lie) are slightly inferior to real-life human motions, with 46.2\% being preferred. Meanwhile ours (GLMI-M) is almost indistinguishable to the real motions. The results suggest the potentials of applying our algorithm to more interesting VR/AR applications. 

\begin{table}[tb]
    \centering
    \begin{tabular}{c|c}
    \toprule
    Preference & Percentage\\
    \midrule
        \textbf{Ours (GLMI-M)} Over \textit{Ours (w/ Lie)} & 0.544 \\
        \textbf{Ours (w/ Lie)} Over \textit{Real Motions} & 0.462\\
        \textbf{Ours (GLMI-M)} Over \textit{Real Motions} & 0.501 \\
    \bottomrule
    \end{tabular}
    \caption{Crowd-sourced subjective assessment to compare motions sampled from \textbf{Ours (GLMI-M)}, \textbf{Ours (w/ Lie)}, and real motions. }
    \label{tab:rank_over}
\end{table}

\begin{figure}[tb]
    \includegraphics[width=0.9\linewidth]{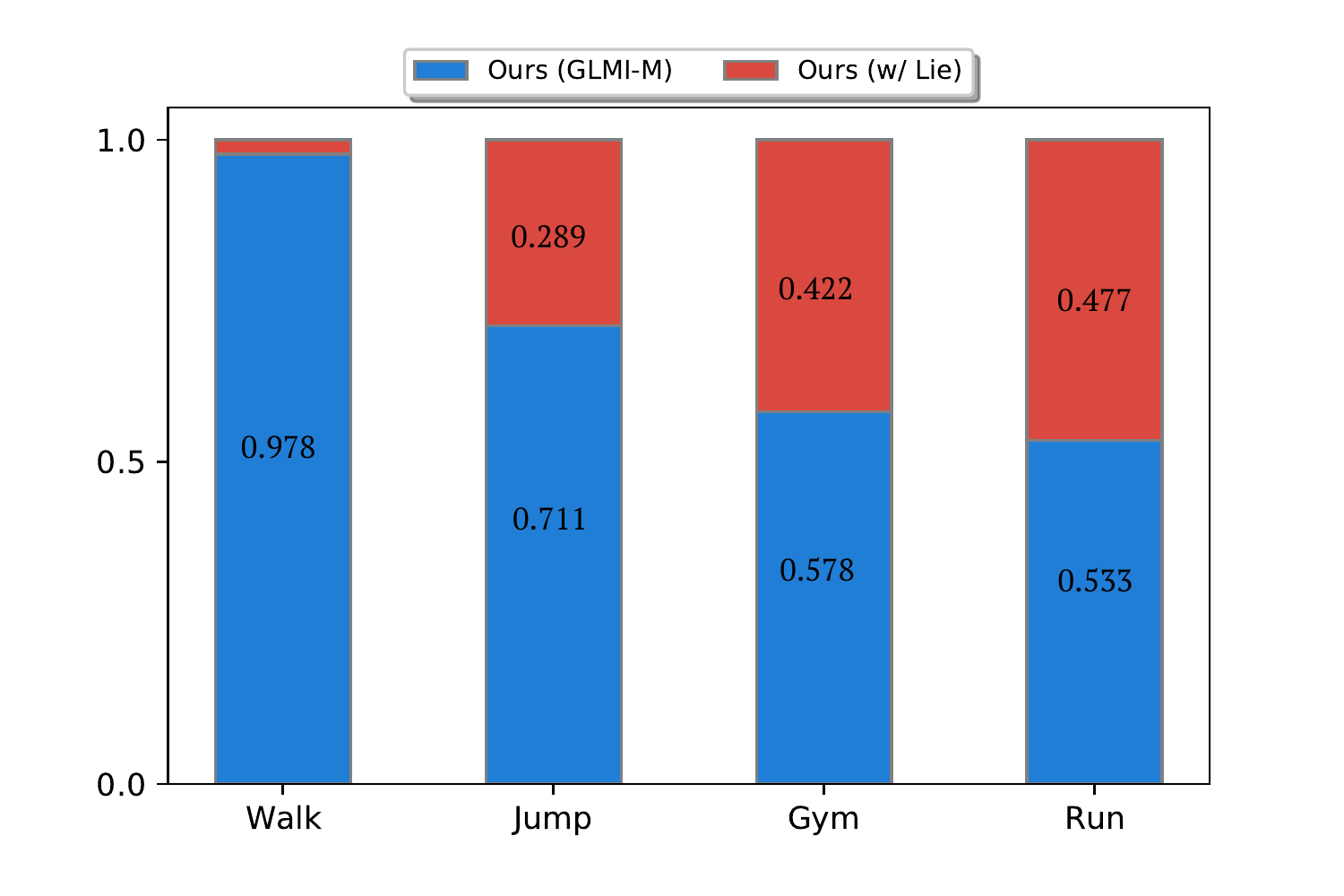}
    \caption{Crowd-sourced subjective assessment to compare generated motions together with their global displacements from \textbf{Ours (GLMI-M)} and \textbf{Ours (w/ Lie)}.}
    \label{fig:rank_locomotion}
\end{figure}

We further investigate the global displacement aspect of the generated motions. As demonstrated in Fig.~\ref{fig:rank_locomotion}, motions generated from ours (GLMI) are always more preferred by users than those from ours (w/ Lie) over all these four action categories. 

In summary, our GLMI-M variant, i.e. ours (GLMI), delivers overall best results among our four action2motion variants, which are often indistinguishable from real-life human motions.


\subsubsection{Locomotion Generation Analysis}
\label{locomotion}

\begin{figure}[tb]

    \begin{tabular}{c}
         \animategraphics[width=0.8\linewidth]{6}{figures/locomotions/Walk5_3d_}{5}{70}  \\
         \animategraphics[width=0.99\linewidth]{6}{figures/locomotions/Walk2_7_3ds_}{5}{70} 
    \end{tabular}
    \caption{Examples of locomotion generated without GLMI (top) vs. with GLMI (bottom). Note the \textit{ghosting} manoeuvre patterns when without GLMI. Best viewed in Adobe Acrobat Reader to see the animations upon clicking.}
    \label{fig:locomotion}
\end{figure}
\begin{table*}[t]
	\centering
	\caption{Performance evaluation over CMU MoCap dataset on two locomotion action types. $\pm$ indicates 95\% confidence interval. $\uparrow$ (or $\downarrow$) is higher (or lower) the better;  $\rightarrow$ means closer to real motion scores the better. For performance, Bold face specifies the best method, with underscore referring to the second best.}
	\label{tab:locomotion}
	\begin{tabular}{l c c c c c c c}
		\toprule
		\multirow{2}{*}{Methods} & \multicolumn{3}{c}{Walk} & & \multicolumn{3}{c}{Jump Forward} \\
		\cline{2-4}
		\cline{6-8}
		& FID$\downarrow$ & Accuracy$\uparrow$ & Diversity$\rightarrow$ &  & FID$\downarrow$ & Accuracy$\uparrow$ & Diversity$\rightarrow$\\   
		\midrule
		
		\textbf{Real motions}    & \et{0.148}{.007}  &  \et{0.999}{.001}& \et{2.618}{.013}& & \et{0.135}{.006}  &  \et{0.999}{.001}& \et{2.711}{.015}\\
		\midrule
		\textit{Action2Motion} (plain)    & \et{6.659}{.119}  &  \et{0.755}{.002}& \et{4.379}{.026} & & \et{13.14}{.104}  &  \et{0.226}{.004}& \et{5.412}{.018}\\
		\textit{Action2Motion} (w/ Lie)   & \et{5.392}{.069}  &  \et{0.786}{.003}& \et{4.200}{.031}  & & \et{7.233}{.124}  &  \et{0.523}{.004}& \et{5.398}{.018}  \\
		\textit{Action2Motion} (GLMI-R)     & \ets{2.096}{.057}  &  \ets{0.930}{.002}& \ets{3.471}{.020}& & \etb{3.796}{.083}  &  \etb{0.749}{.018}& \etb{4.662}{.031} \\
		\textit{Action2Motion} (GLMI-M)   & \etb{1.183}{.028}  &  \etb{0.967}{.001}& \etb{3.059}{.022} & & \ets{4.443}{.146}  &  \ets{0.715}{.005}& \ets{4.747}{.031} \\
		\bottomrule
	\end{tabular}
	\vspace{-0.2cm}
\end{table*}

\begin{figure}[tb]
	\centering
	\begin{tabular}{c}
	    \underline{\tiny{\textbf{Click}$\downarrow$}} \\
	    \animategraphics[height=0.95cm]{6}{figures/interpolation/lift_dumbbell_}{0}{59}    \\
	     \includegraphics[width=\linewidth]{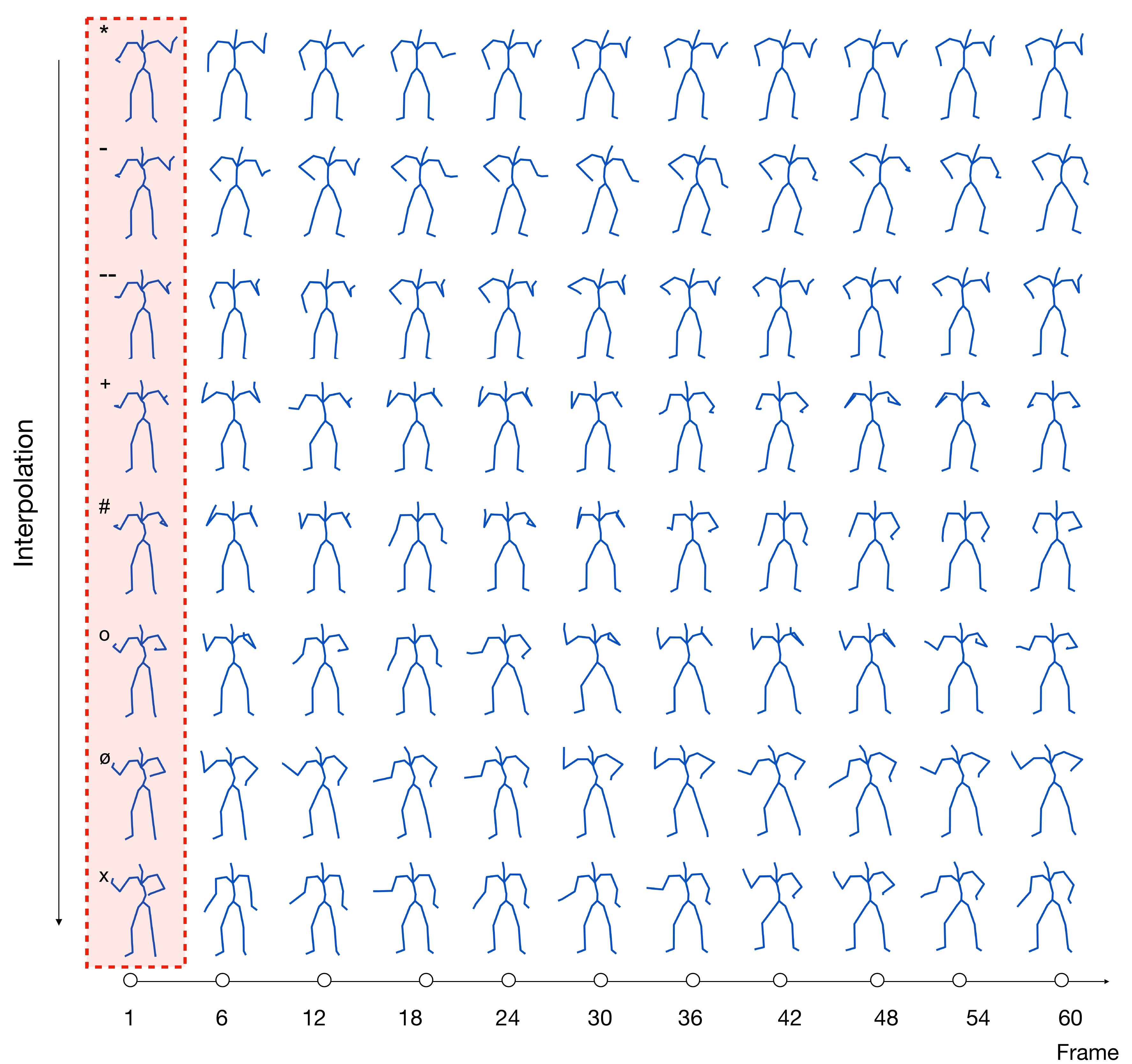} 
	\end{tabular}
	\caption{Examples of motion interpolation in \textit{lift dumbbell}. Every 6th frame is shown. See text for details.
	Best viewed in Adobe Acrobat Reader to activate the animations by clicking the boxed items in the top row. Note each item in the top row is with specific tag corresponding to its row of motion sequence displayed below.
	}
	\label{fig:interpolation}
\end{figure}

Locomotions (e.g. walking) are the most common activities in our daily life, which typically involve full-body displacements. Fig.~\ref{fig:locomotion} visually compares walking motions produced with vs. without our global local movement integration (GLMI) module. When without, the walking motions appear surreal like \textit{ghost} haunting on the ground, with arm and leg local movements not tuned to its global motion trajectory. By contrast, our proposed GLMI module significantly mitigates these issues. For example, the waving patterns of left (or right) arm is now synchronized with the right (or left) leg; the local-part moments are also well in agreement with the full-body motion trajectories.

Table~\ref{tab:locomotion} quantitatively evaluates the effects of incorporating GLMI module for locomotion generation on CMU MoCap dataset. The same evaluation metrics of Section~\ref{subsubsec:metrics} are considered here. The number of motion sampling is set to 500. 
Overall, ours with GLMI variants perform best over all the three metrics. In contrast, ours (plain) attains worst results, which we attribute to the missing modules of Lie algebraic representation and GLMI. 
Moreover, GLMI-M , i.e. GLMI with MLP implementation, works best in generating \textit{Walking} motions, while GLMI-R takes the lead in \textit{Jump Forward}.


\subsubsection{Interpolation in Latent Space}
\label{interpolation}

\begin{figure*}[t]
	\centering
	\begin{tabular}{c}
	    \underline{\tiny{\textbf{Click}$\downarrow$}}\\
	     \animategraphics[height=1.2cm]{6}{figures/action_transition/transition_}{0}{99}  \\
	     \includegraphics[width=0.99\linewidth]{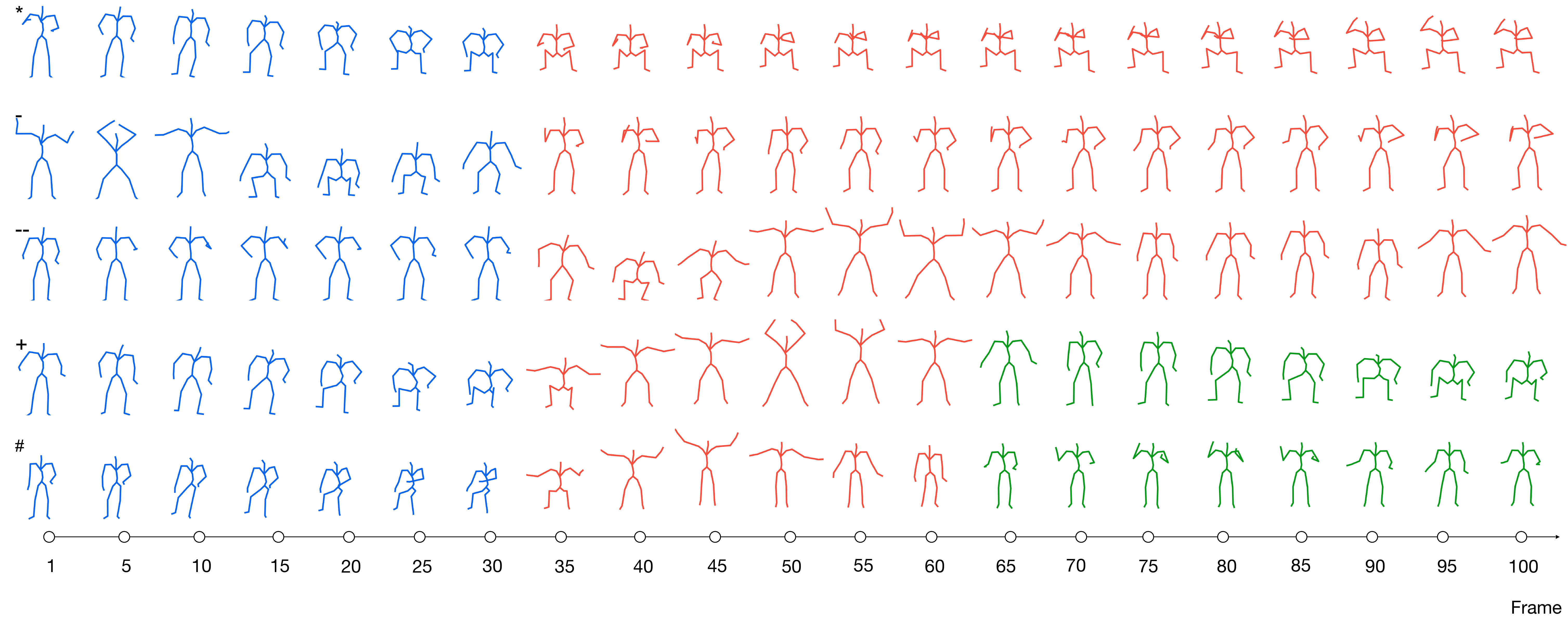} 
	\end{tabular}
	\caption{Action transition examples. Every 5th frame is shown. The top three rows show transition between two actions. from top to bottom, they are \textit{sit-drink}, \textit{jump up-lift dumbbell}, \textit{lift dumbbell-jump up}, respectively. 
	The bottom two rows display transition of three actions, which are (from top to bottom) \textit{sit-jump up-sit} and \textit{sit-jump up-lift dumbbell}, respectively. 
	Best viewed in Adobe Acrobat Reader to activate the animations by clicking items in the top row. Note each item in the top row is with specific tag corresponding to its row of motion sequence displayed below.
	}
	\label{fig:action transition}
\end{figure*}
\begin{figure}[t]
	\centering
	\begin{tabular}{c}
	     \animategraphics[width=\linewidth]{6}{figures/inpainting/walk_mul_}{0}{100}  \\
	\end{tabular}
	\caption{Examples of motion \revision{outpainting} of \textit{Walking}. Provided several initial poses (in black), our method completes the rest motion sequence with multiple plausible outcomes. Best viewed in Adobe Acrobat Reader to see the animations upon clicking.}
	\label{fig:outpainting}
	\vspace{-0.3cm}
\end{figure}

Generative models could be regarded as a function mapping between points in a latent space and those in the real data space. 
Meanwhile, similar to the concept of well-posed problems, a well-learned generative model is expected to behave smoothly from a small perturbation in the latent space. 
In other words, when we perform interpolations between two distinct latent codes, their generated motions are supposed to transit smoothly. 
It is thus of interest to examine how interpolations in the latent space would change the motion generation behaviors of our action2motion. It also demonstrates the model capability in producing non-existent samples.

The task is a bit complicated in our situation, as our model generates motion sequences instead of single images. 
Alternatively, we use the first poses as anchors to perform interpolation between two motions.  Specifically, the first poses of two pose sequences are selected. Then, a series of points can be created on the linear path between the latent vectors (i.e. noise vectors) of these two poses. After that, these points are input as initial latent vectors into our model to kick-start the generation of rest poses. 

Fig.~\ref{fig:interpolation} considers \textit{lift dumbbell} action.  
Here two pose sequences are deliberately selected from motions generated by action2motion (GLMI-M), where the first poses of the two sequences are a person lifting with the left (and the right) hand, respectively. 
We have the following observations. 1) As demonstrated in the first column, transition from the \textit{left hand} pose to the \textit{right hand} pose is realistic at the first poses, by gradually putting one hand down and lifting another hand up. 2) From each of these initial interpolated poses, a visually natural motion sequence is generated. 3) Interestingly the interpolation leads to the generation of a novel motion, \textit{lift dumbbell with both hands}.


\subsubsection{Action Transition}
\label{transition}
To showcase the flexibility of our motion synthesis process, \textit{action transition} is explored by switching the action categories during sequence generation. Exemplar results are presented in Fig.~\ref{fig:action transition}. 
To our surprise, our action2motion model is able to produce unseen motions through action transition. 
In the first row of Fig.~\ref{fig:action transition}, after switching from \textit{sit down} to \textit{drink}, the character starts to open the bottle and drink with a sitting pose. However, all drinking motions in our training set are performed in \textit{standing} poses. 
As shown in these examples, the resulting motion sequences are rather realistic and with natural transitions which is well maintained in transitions of not only two actions, but also three actions. 
This experiment clearly demonstrates the capacity of our approach in synthesizing unseen motions that goes beyond merely memorizing training examples.


\begin{figure*}[thb]
    \centering
	\includegraphics[width=0.95\linewidth]{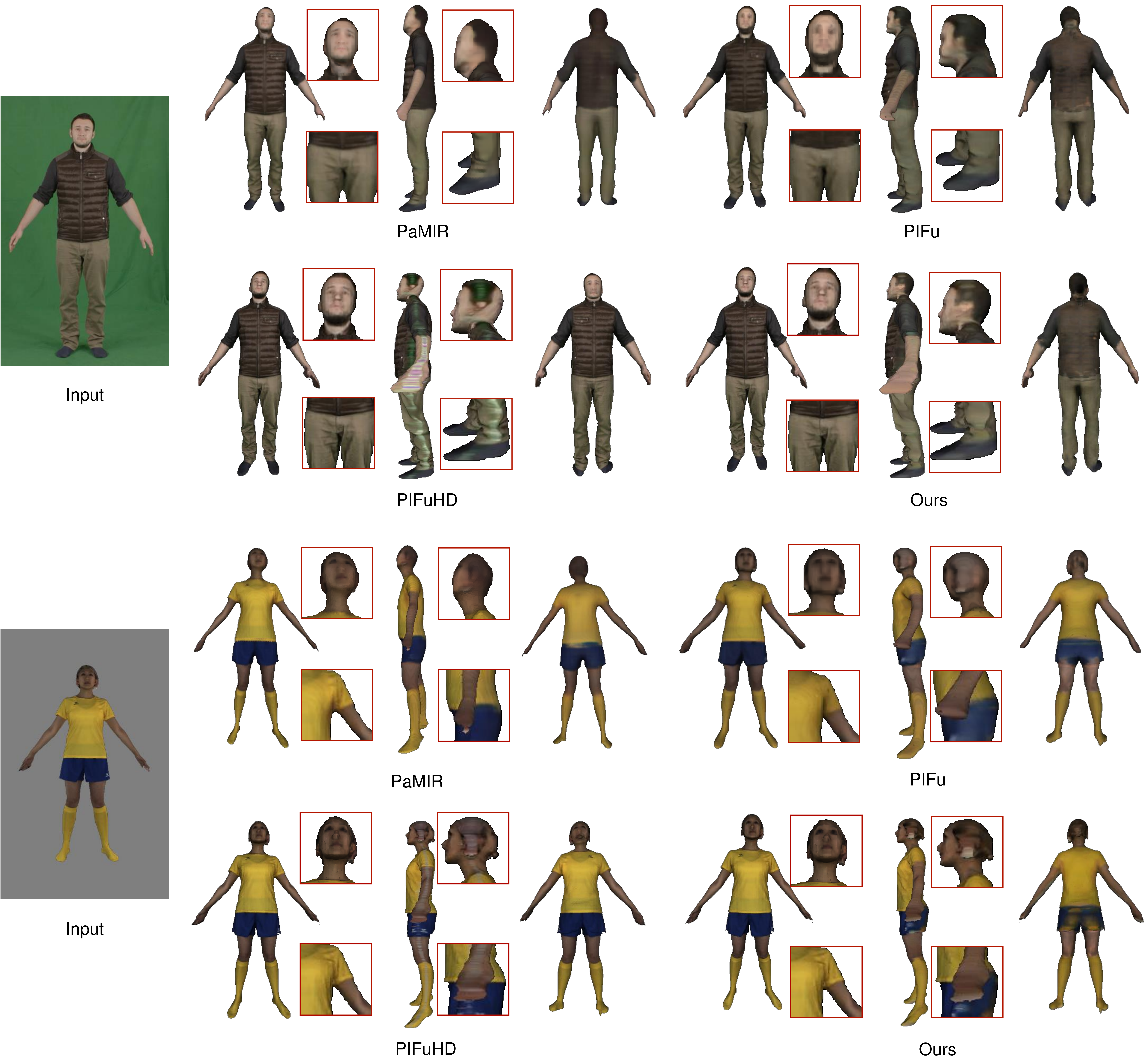}
	\caption{\revision{A qualitatively comparison of reconstructing 3D human shape \& texture from single image.
	The input single images are show on the left, where the top image is from People Snapshot dataset~\citep{alldieck2018video} and the bottom one from BUFF dataset~\citep{Zhang_2017_CVPR}.
	Comparing with the state-of-the-art methods of PaMIR~\citep{zheng2021pamir}, PIFu~\citep{saito2019pifu} and PIFuHD~\citep{saito2020pifuhd}, 
	our approach improves upon PIFuHD and PIFu by integrating their otherwise segregated strengths of high-resolution geometry and high-quality texture at novel views.}}
	\label{fig:3d_mesh_recovery}
\end{figure*}

\begin{figure*}[thb]
    \centering
	\includegraphics[width=0.95\linewidth]{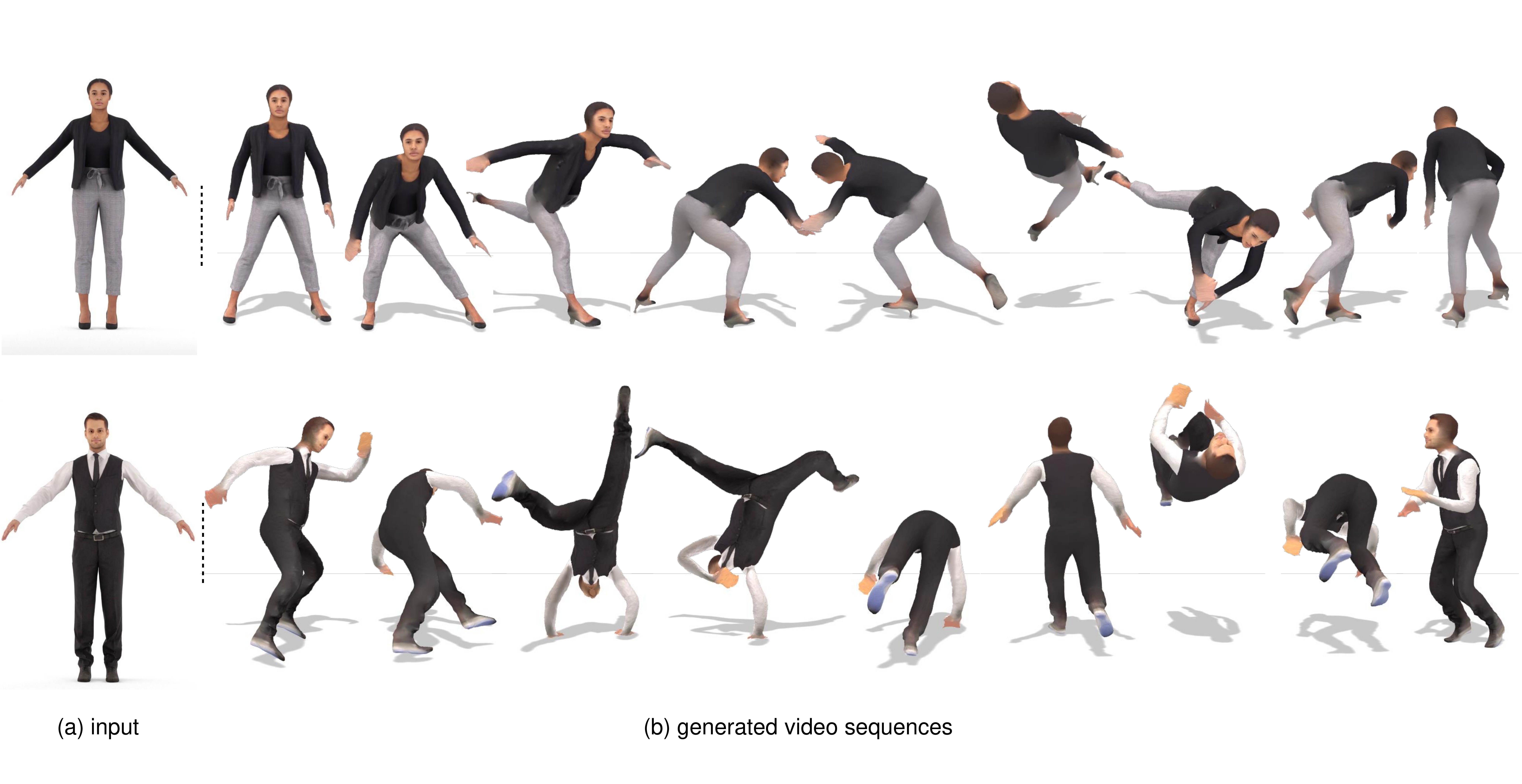}
	\caption{\revision{Two animation results of our method. Given single images of frontal view of individuals shown on the left,  their 3D shapes are reconstructed, 2D videos are obtained, using prescribed off-the shelf motion sequences. The videos produced by our method are visually plausible.}}
	\label{fig:video_animation}
\end{figure*}

\subsubsection{\revision{Motion Outpainting}}
\label{outpainting}

Our method could also serve as a motion \revision{outpainting} tool: provided the initial few poses, apply our method to complete the rest of the motion sequence. 
This is realized by simply fixing the beginning poses, and generating the rest. Executing multiple independent runs usually creates distinct yet plausible outcomes. 
Fig.~\ref{fig:outpainting} illustrates such an example. Here black poses denote the fixed initial poses of \textit{Walk}. This is completed by our model with visually plausible walking motions of distinct velocities and directions. 
This also suggests the necessity of modeling motion forecasting and generation in a non-deterministic manner.

\subsection{Step 2: Motion2video} 
\label{subsec:motion2video}
\revision{Side-by-side evaluations are performed in terms of reconstructing 3D human shapes \& textures from single images in Sec.~\ref{subsubsec:3d_shape_reconstruc}, and animation in Sec.~\ref{subsubsec:motion2video}.}

\subsubsection{3D Shape and Texture Reconstruction}
\label{subsubsec:3d_shape_reconstruc}

\begin{table}[tb]
    \centering
    \begin{tabular}{c|c}
    \toprule
    Method & Average Rank$\downarrow$\\
    \midrule
        PIFuHD \citep{saito2020pifuhd} & 3.60 \\
        PIFu \citep{saito2019pifu} & 2.36\\
        PaMIR \citep{zheng2021pamir} & 2.26\\
        \textbf{Ours} & 1.77\\
    \bottomrule
    \end{tabular}
    \caption{\revision{Quantitative comparison of reconstructing 3D human shape \& texture from single images. The numbers are averaged user preference ranks, with $\downarrow$ meaning the numbers are lower the better.}}
    \label{tab:3d_shape_average_rank}
\end{table}

\begin{figure}[tb]
  \centering
  \includegraphics[width=\linewidth]{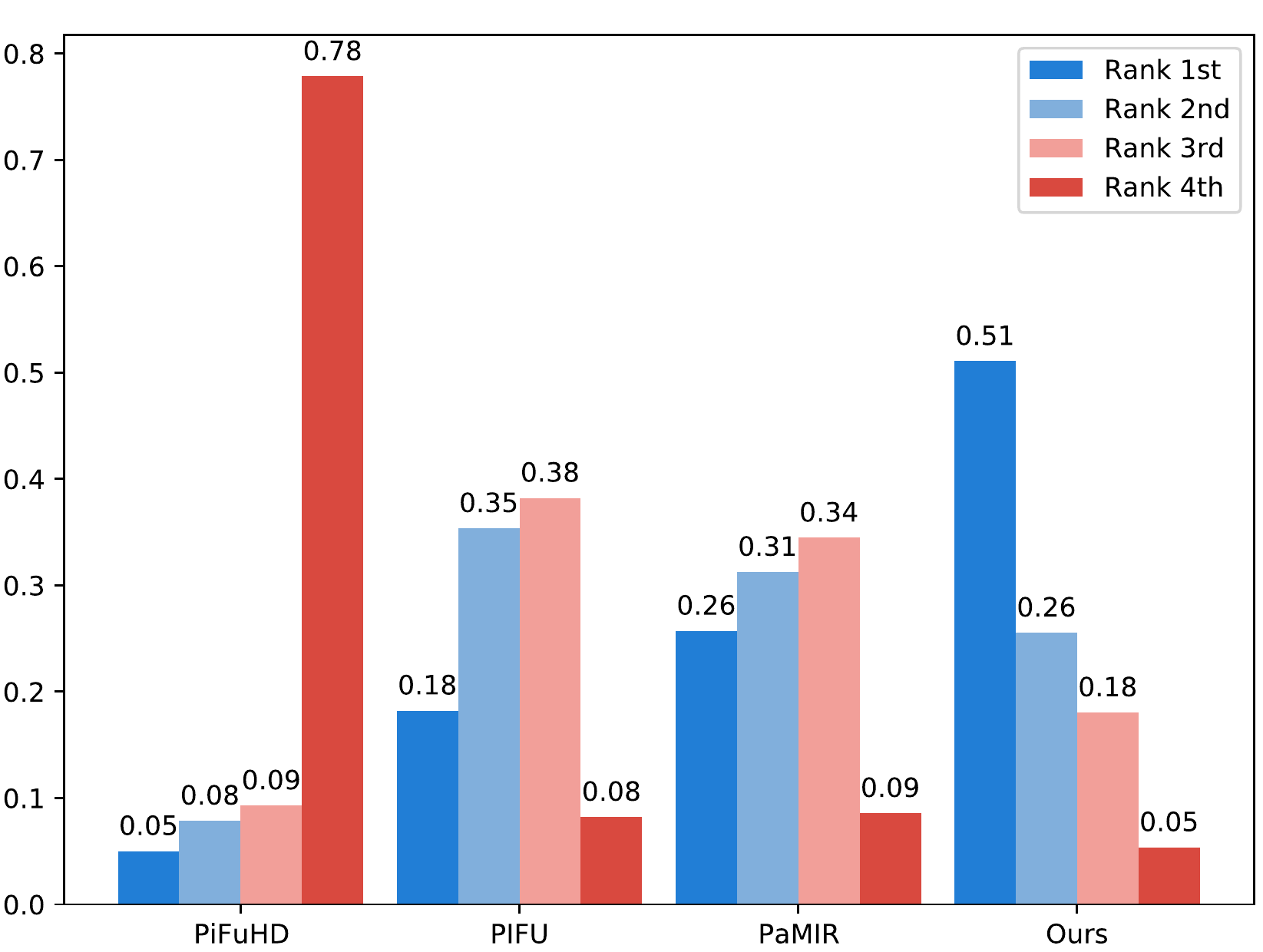}
  \setlength{\abovecaptionskip}{0.1cm}
  \setlength{\belowcaptionskip}{-0.2cm}  
  \vspace{-7pt}
  \caption{\revision{User preference distributions of reconstructing 3D human shape \& texture from single images.}}
  \label{fig:rank_hist_shape}
  \vspace{-0.3cm}
\end{figure}

\revision{Here we focus on the evaluation of reconstructing 3D human shape \& texture from single images, where the respective part of our approach is compared side-by-side with the state-of-the-arts, namely PaMIR~\citep{zheng2021pamir}, PIFu~\citep{saito2019pifu} and PIFuHD~\citep{saito2020pifuhd}. 
PaMIR~\citep{zheng2021pamir} combines parametric SMPL body model with deep implicit function for robust 3D shape reconstruction. In our comparison, 30 images are obtained from a wide variety of sources, including the BUFF dataset~\citep{Zhang_2017_CVPR}, the People Snapshot dataset~\citep{alldieck2018video}, internet images, CG image, and our in-house captured images. 
Following the network architectures, the input resolution of PaMIR and PIFU is $512\times512$, whereas the input image resolution is $1024\times1024$ for PIFuHD and our approach.}

\revision{Exemplar results of reconstructed textured shapes from single input images are shown in Fig.~\ref{fig:3d_mesh_recovery}. The shapes and textures extracted by PaMIR and PIFu commonly lack details, 
and are oftentimes inaccurate. For example, the 3D shape of lady produced by PaMIR is overly slim, together with an smooth face that lack geometric details which is noticeable especially from side-views. 
PIFuHD is capable of recovering 3D shapes with better facial geometry and in high-resolution, yet the texture is often visually unpleasantly wrong, especially when viewing from the back. 
In contrast, our method maintains a delicate balance of shape and texture, thus stands at a better position in facilitating the follow-up animation and realistic rendering processes in our pipeline.}

\revision{For quantitative evaluation, user study is further conducted to measure the perceptual quality of the comparison methods. 
For each input image, 20 Amazon mechanical turk Workers are enrolled to rank their preferences over the shapes reconstructed by their corresponding comparison methods. 
Table~\ref{tab:3d_shape_average_rank} displays the average rank of each method, with more detailed rank distributions presented in Fig.~\ref{fig:rank_hist_shape}. 
Our method clearly stands out with the most appreciations from users, where almost half (i.e. 51\%) results are ranked the first. 
By contract, PIFuHD is the least preferred one, of which 78\% results are placed as least favorable. 
In-between are PIFu with the second lowest average rank, and PaMIR that receives considerable more positive feedback compared to PIFuHD.}

\subsubsection{Motion2Video Animation}
\label{subsubsec:motion2video}

\begin{figure*}[thb]
    \centering
	\includegraphics[width=0.95\linewidth]{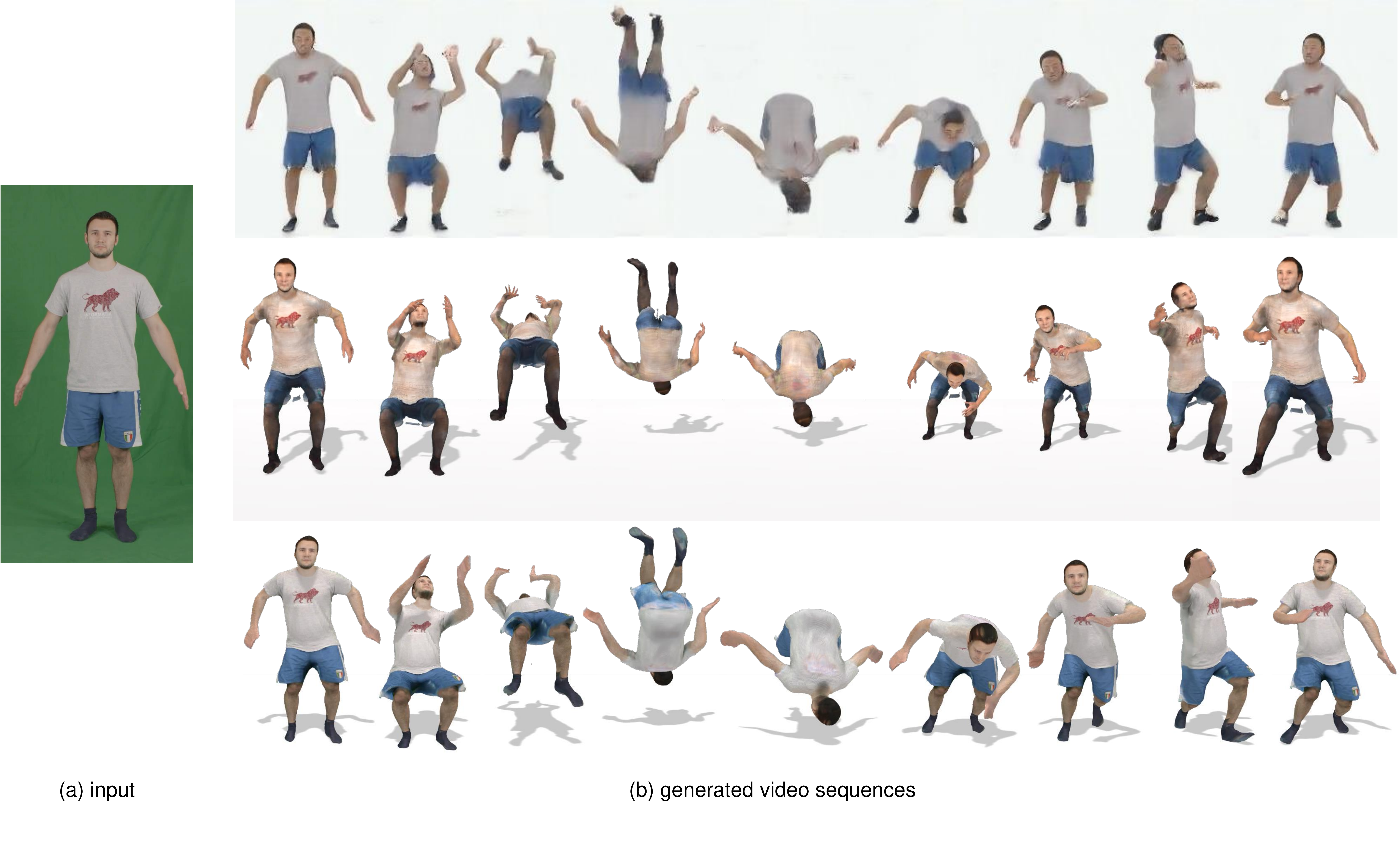}
	\caption{\revision{Comparing our method (bottom) with Liquid Warping GAN~\citep{liu2019liquid} (top) and ARCH~\citep{huang2020arch} (middle), animated using the same input image and motion sequence. Results are displayed by pairing the corresponding video frames.}}
	\label{fig:liquidwgan_com}
\end{figure*}

\begin{figure*}[thb]
	\centering
	\includegraphics[width=0.95\linewidth]{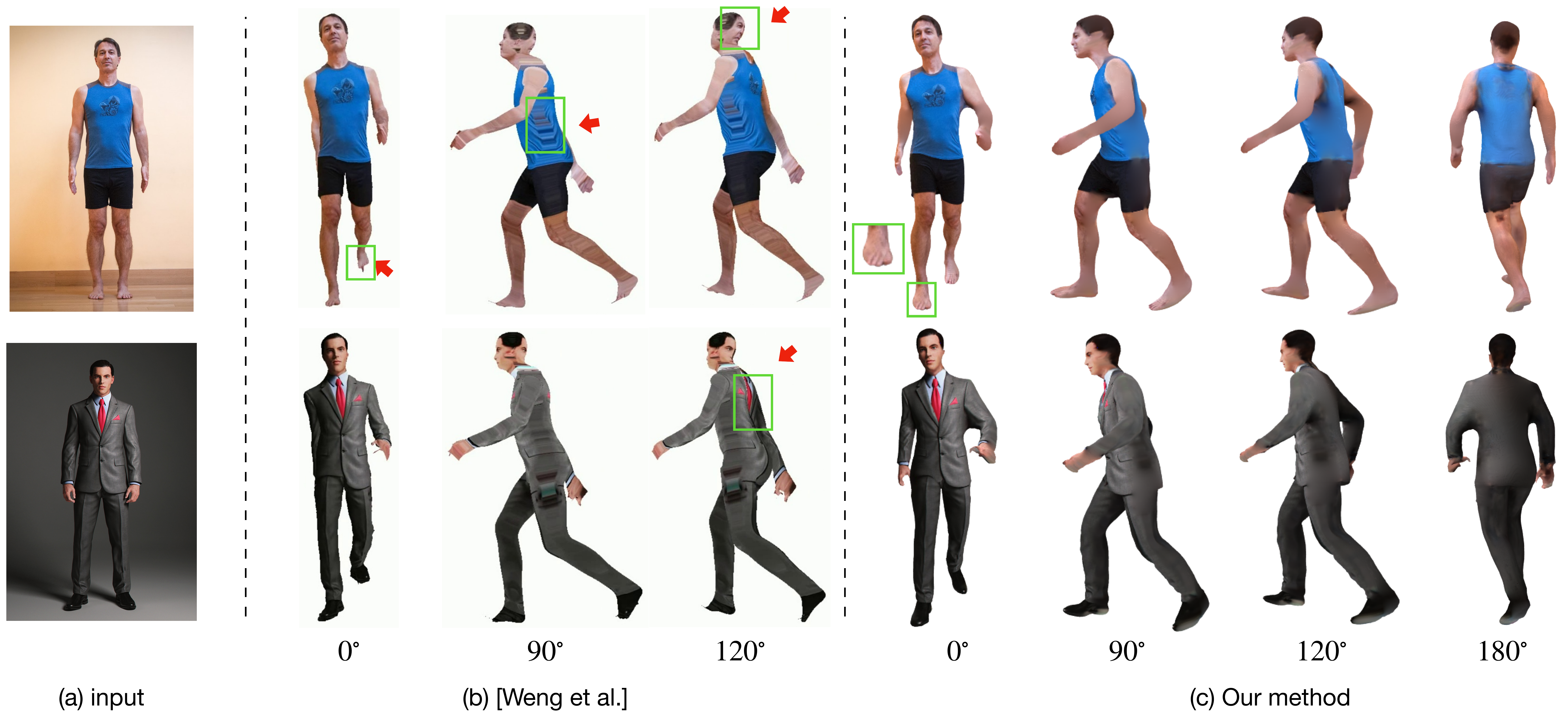}
	\caption{Comparing our action2video with~\cite{weng2019photo} by animating walking motions. For each given image on the left, we show the results of~\cite{weng2019photo} (middle column) and ours (right column) from different views. \cite{weng2019photo} fail to build an intact 3D texture model (e.g. incomplete feet), and the appearance of unseen part is distorted. Our method could generate plausible animation from all angles.}
	\label{fig:video_comparison}
	\vspace{-0.3cm}
\end{figure*}

\begin{figure}[thb]
    \centering
	\includegraphics[width=0.9\linewidth]{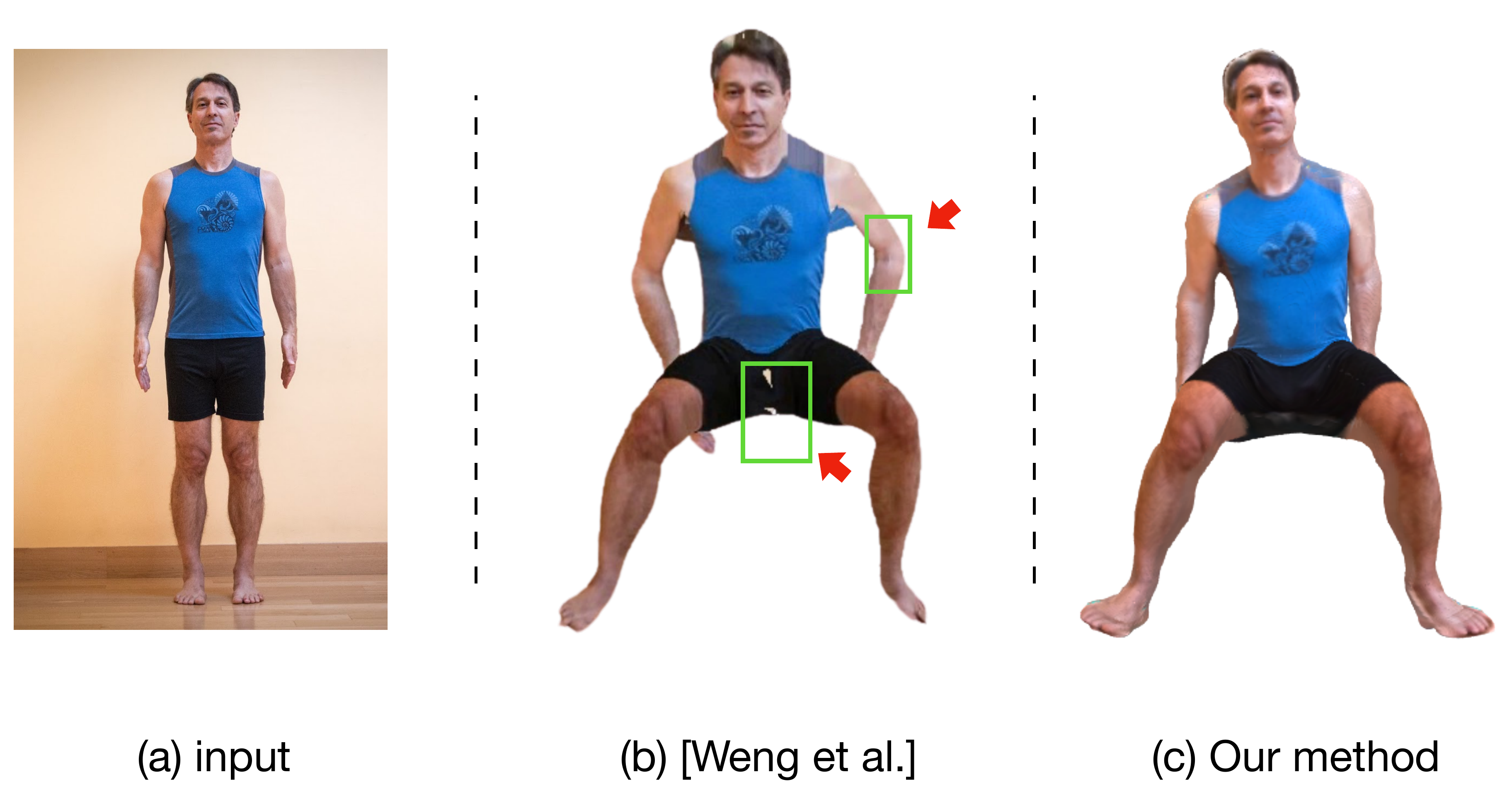}
	\caption{Comparing our method with~\cite{weng2019photo} by animating sitting motions.}
	\label{fig:video_comparison2}
\end{figure}

\begin{table}[tb]
    \centering
    \begin{tabular}{c|c}
    \toprule
    Preference & Percentage\\
    \midrule
        \textbf{Ours} Over \citep{liu2019liquid} & 0.843 \\
        \textbf{Ours} Over \citep{huang2020arch} & 0.593 \\
        \textbf{Ours} Over \citep{weng2019photo} & 0.703 \\
    \bottomrule
    \end{tabular}
    \caption{\revision{Crowd-sourced subjective assessment to compare the videos animated with the same image and motion, produced by \textbf{Ours}, \textbf{Liquid Warping GAN}~\citep{liu2019liquid}, \textbf{ARCH}~\citep{huang2020arch} and \textbf{~\cite{weng2019photo}}. }}
    \label{tab:video_animation_prefer}
\end{table}

\revision{In Fig.~\ref{fig:video_animation}, We present two single image animation showcases using our method. 3D shape and texture are predicted from input images, which are driven by two challenging motions, cartwheel, from Adobe Mixamo~\footnote{www.mixamo.com}. As shown, our method could obtain accurate shape and texture predictions from all views, as well as plausible animations with provided motions.} 

\revision{In what follows, we elaborate the comparisons between our method and other three state-of-the-art image animation methods~\citep{liu2019liquid,huang2020arch,weng2019photo}. For quantitative evaluation, we conduct user study on Mechanical Turk which pairs the videos animated with the same image and motion from our and comparison method, and request the workers to determine which one that is "more realistic". For each animation, 50 workers with Hit approval rate higher than 97\% are enrolled for perceptual assessment. }

\textbf{Comparison with Liquid Warping GAN~\citep{liu2019liquid}}. \revision{Liquid Warping GAN~\citep{liu2019liquid} is a learning based motion transfer method in pseudo-3D space, where 3D SMPL model estimated from reference video frames are used to re-pose the person in source image. 
Fig.~\ref{fig:liquidwgan_com} presents the animated videos by our method (bottom) and Liquid Warping GAN (top), when feeding with the same input image and motion. While successfully modeling the motion dynamics,  the individual images obtained by Liquid Warping GAN are very blurring such that the characteristic personal landmarks of face or T-shirt logo are nearly unrecognizable. In contrast, the animation results of our method are of high-resolution and high quality. 
A user study is performed for quantitative evaluation, based on 22 animations from Liquid Warping GAN and our method covering a variety of input images and motion sequences, including composed of 9 Mixamo motions and 13 motions generated by our action2motion step. 
As shown in Table~\ref{tab:video_animation_prefer}, 84.3\% of our animations are preferred by users.}

\begin{figure}[thb]
    \centering
	\includegraphics[width=0.9\linewidth]{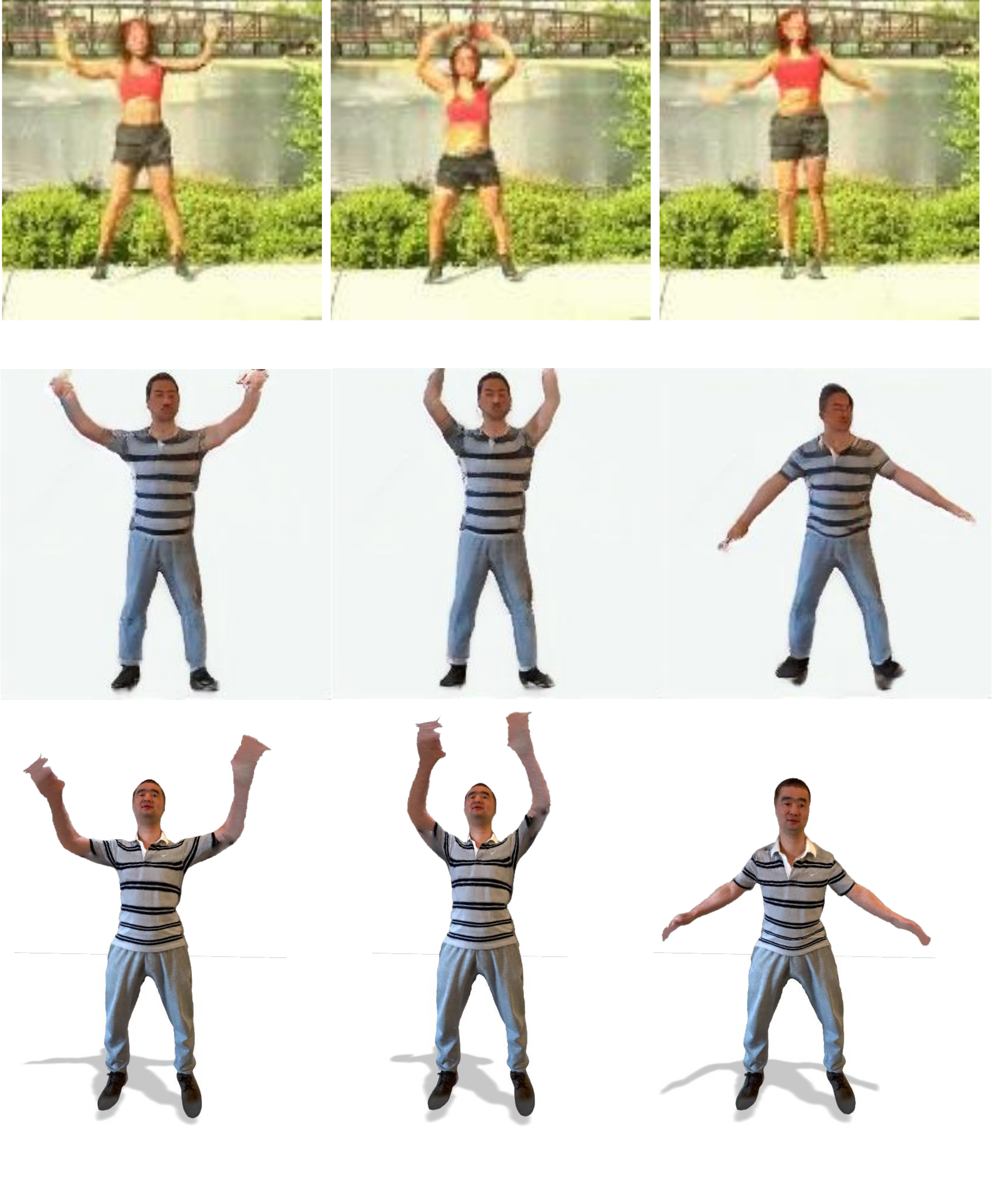}
	\caption{\revision{Visual comparison of three methods: (top) a state-of-the-art 2D-based method~\cite{kim2019unsupervised}, (middle) Liquid Warping GAN~\citep{liu2019liquid}, and (bottom) ours.}}
	\label{fig:video_2d_comparison}
\end{figure}

\begin{figure*}[thb]
	\centering
    \includegraphics[width=0.89\linewidth]{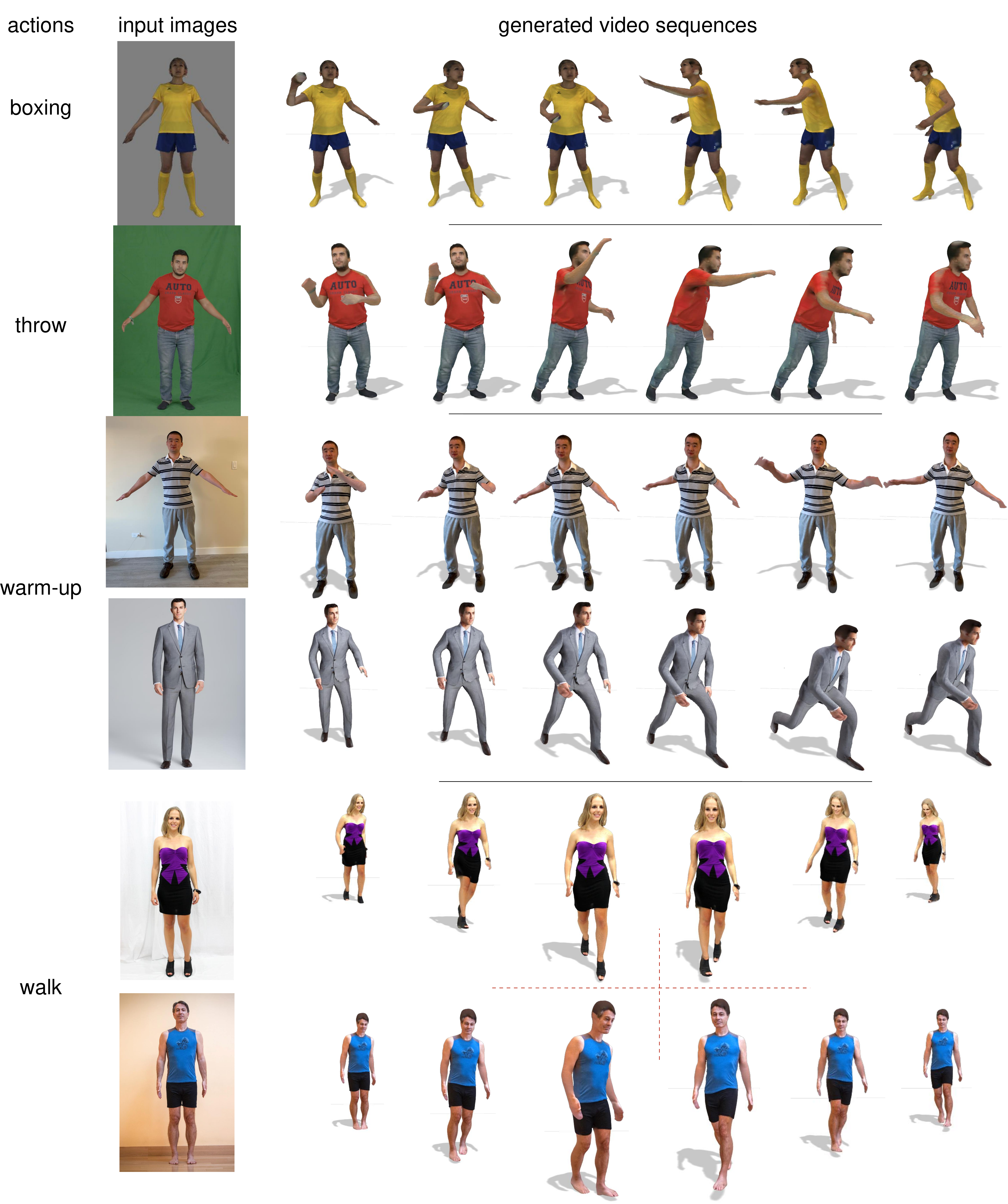}	
    \caption{Exemplar videos produced by our action2video pipeline. Given a reference image and a specific action category, our action2video could extract 3D human shapes \& cloth textures, and animate \& render into diverse motion videos. For boxing and throw actions, one video are shown, each animated by a different 3D character extracted from a single image; similarly, two distinct videos and four distinct videos are presented for the warm-up action and walking action respectively.}
	\label{fig:video_diverse}
	\vspace{-0.3cm}
\end{figure*}

\begin{figure*}[thb]
	\centering
    \includegraphics[width=0.9\linewidth]{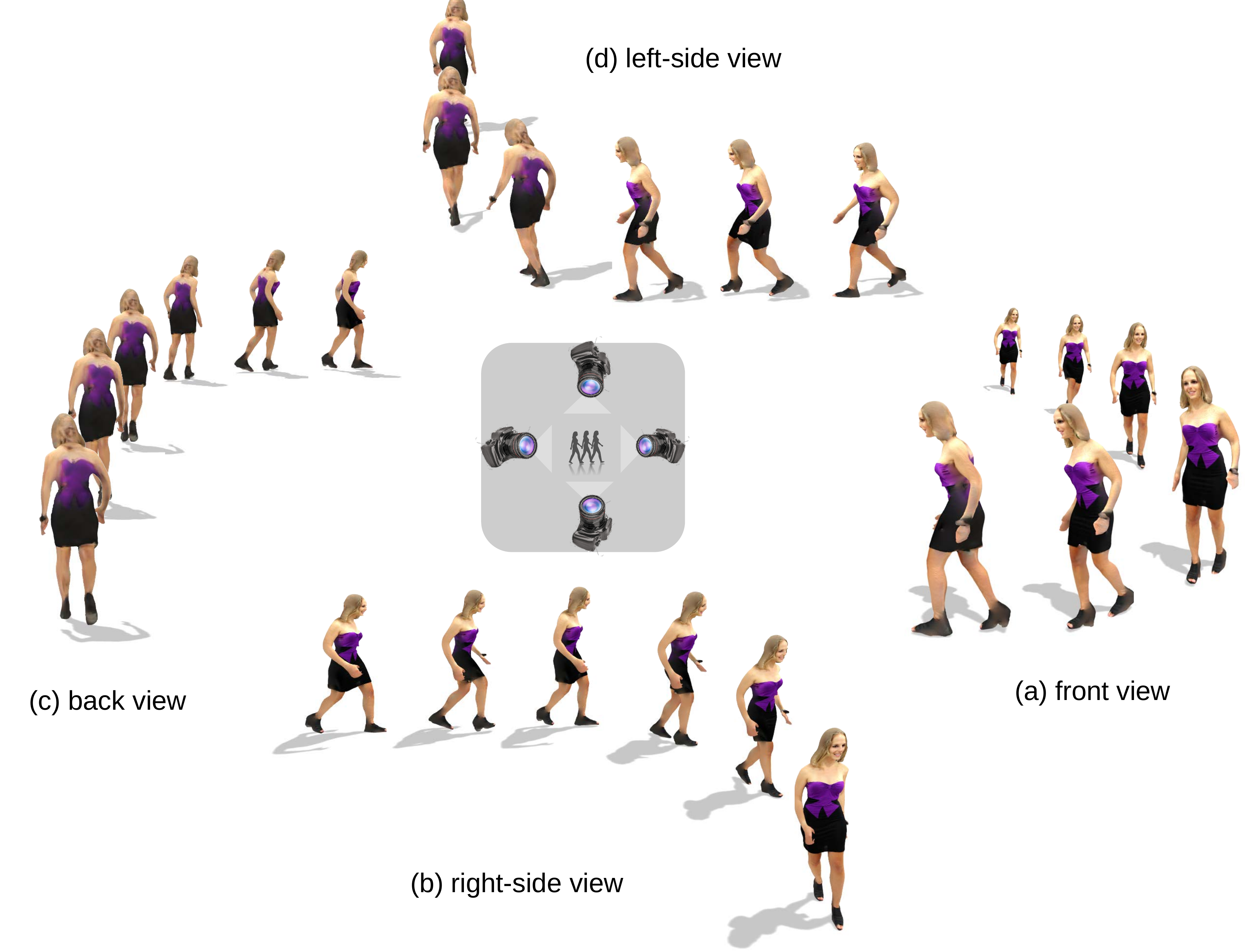}	
    \caption{An generated walking video from the following views: (a) front, (b) right-side, (c) back, and (d) left-side.}
	\label{fig:four_cameras}
	\vspace{-0.3cm}
\end{figure*}

\textbf{Comparison with ARCH~\citep{huang2020arch}.} \revision{ARCH~\cite{huang2020arch} uses a semantic deformation field to produce 3D rigged full-body human avatars from a single image, which is already animatable. However, the implementation and pre-trained model of ARCH has not been released yet. We managed to obtain 3 animated 3d model sequences from the authors with our provided images and Mixamo motions. We render video frames of these 3d model sequences in Unity3D with the same environment setting (e.g. light, camera) as ours. Fig.~\ref{fig:liquidwgan_com} presents a visual comparison between ARCH's result (middle) and our result (bottom). Though ARCH shows capability of generating reasonable rendering, the person appearance is yet to be realistic. For example, the pants comes with several blue debris; the two feet of the man are in wrong color (black); and the texture of T-shirt is overly bright. A user study is again conducted regarding the 3 animations from ARCH and our method. As given in Table~\ref{tab:video_animation_prefer}, our method earns more preference (i.e. 59.3\%) from users. Please refer to the supplementary video for more visual comparisons.}

\textbf{Comparison with~\cite{weng2019photo}}. the work of~\cite{weng2019photo} is also closely related to part of our motion2video step, where a 3D character is extracted out of a single image and is further animated to form videos. 
Their implementation is unfortunately not publicly available, instead we obtain from the authors of~\cite{weng2019photo} two animated action sequences (i.e. sit and walk) from the two input images provided by us. Note that the motions involved in~\cite{weng2019photo} are \textit{real} MoCap motion sequences, while our motions are \textit{generated} by ourselves. 
For an easy side-by-side visual comparison, we hand pick two of our generated motions that resemble the animations used~\citep{weng2019photo}. 
The \textit{walk} and \textit{sit} visual results are displayed and compared in Figs.~\ref{fig:video_comparison} and~\ref{fig:video_comparison2}, respectively.

When viewing from frontal view, the results of~\cite{weng2019photo} possess incomplete and distorted errors including the incomplete feet (Fig.~\ref{fig:video_comparison}(b)), over-slim arms, and torn pants (Fig.~\ref{fig:video_comparison2}(b)), as highlighted by red arrows. These artifacts come from the fact that the textures are directly copied and pasted from the 2D input image, which is inadequate to maintain intact appearance in 3D geometry. In comparison, our results are noticeably better at preserving detailed structure and appearance, e.g. around the feet. 

When inspecting from the side and back views of the extracted 3D characters that are not directly visible from the input image view, the textured results of~\cite{weng2019photo} are simply mirrored from the frontal region, as shown in the back side of head and torso - the visual results are thus significantly deteriorated to being funny. 
In contrast, our results preserve reasonable 3D shape and consistent appearance across multiple views including the frontal view. 
\revision{Moreover, a similar user study is conducted among the two set of generated videos. As in Table~\ref{tab:video_animation_prefer}, our method is 70.3\% more preferred over \cite{weng2019photo}.}

\subsection{The Full Action2Video Pipeline}
\label{subsec:action2video}

This section is devoted to the examination of our full action2video pipeline. We start by comparing with state-of-the-art 2D-based human video generation results. 
Further experiments also demonstrate the capacity of our action2video approach in accommodating input images from different sources.

\textbf{Comparison with existing methods}. The work of~\cite{kim2019unsupervised} is state-of-art in generating human motion videos, which is 2D-based and relies on large-scale training set of videos. 
Fig.~\ref{fig:video_2d_comparison} presents a comparison of their results and ours that share in common similar poses and views. 
Compared with our results, the frames of~\cite{kim2019unsupervised} is of low resolution (128x128). Moreover, there are visible lack of details of face, hands \& clothes, and unrealistic shape deformations, which we attribute to their innate 2D based limitations. For example, lengths of legs and arms in~\cite{kim2019unsupervised} of the same lady character vary over time. \revision{Moreover, as presented in the middle row of Fig.~\ref{fig:video_2d_comparison}, the exemplar video result generated by engaging Liquid Warping GAN based on the same motion generated by our action2motion step, where edges and facial details are very foggy and fuzzy, when comparing to our results shown at the bottom row.}

\textbf{Diverse input image sources.} This experiment is to evaluate the flexibility of our action2video pipeline in accommodating input images from varied sources. 
Fig.~\ref{fig:video_diverse} presents our action2video results based on BUFF images (e.g. 1st row), People Snapshot images (e.g. 2nd row), Internet images (e.g. 4th row), these captured by our mobile-phone (e.g. 3rd row) as the input images.
Overall our approach is able to adapt to these different applications, and to produce videos of visually pleasing quality. More visual results are shown in the supplementary video.

\textbf{Multiple camera views.} Fig.~\ref{fig:four_cameras} displays an exemplar video sequence generated by our approach, that is inspected from four different views. It demonstrates 1) our extracted 3D shape and clothing texture are reasonably realistic when examined in different rendered views, and 2) compared to the popular 2D-based methods, our generated videos are consistent among distinct views.

\section{Conclusion and Discussion}
\textbf{Conclusion.} We propose an action2video approach to tackle the exciting and challenging problem of generating natural and diverse 3D motions \& videos of human actions. 
This is accomplished in this paper by a 2-step pipeline: action2motion focuses on generating 3D human motions, which are then turned into videos by motion2video. 
Empirical studies demonstrate the effectiveness of our approach.

\noindent{\textbf{Limitation and Future Work.}} Our approach performs reasonably well in practice; empirically it outperforms the state-of-the-art methods in many aspects along the full pipeline. 
On the other hand, we recognize that our training set, primarily the in-house HumanAct12 dataset is relatively small, which contains 1,191 motions.
For future work, we plan to acquire a larger dataset with broader set of actions, to generate motions and videos from a wider range of human activities including interactions with multiple people, with surroundings and objects, \revisions{and to improve the reconstructed shape details such as fingers.} Furthermore, we would investigate its possible applications such as augmenting data for human-centric tasks (action recognition, pose estimation), and VR/AR.

\begin{acknowledgements}
This work is partly supported by the NSERC Discovery Grant and the UAHJIC grants. We would like to thank the Amazon MTurk workers for their efforts in user studies, and the anonymous reviewers for helping to improve the presentation of our manuscript. We also want to thank Chungyi Weng, Yuanlu Xu and Zerong Zheng for their great help on reproducing the works of Photo Wake-Up, ARCH and PaMIR.
\end{acknowledgements}

%
%

\newpage
\newpage
\typeout{}
\bibliographystyle{spbasic}      
\bibliography{sample}   


\end{document}